\documentclass[twoside,11pt]{article}

\usepackage[preprint]{jmlr2e}

\usepackage{amsmath}
\usepackage{float}
\usepackage{adjustbox}
\usepackage{booktabs}
\usepackage{xcolor}
\usepackage{algpseudocode}
\usepackage{lastpage}

\hypersetup{
    colorlinks,
    linkcolor={red!50!black},
    citecolor={blue!50!black},
    urlcolor={blue!80!black}
}

\newenvironment{thm}{\begin{theorem}}{\end{theorem}}
\newenvironment{cor}{\begin{corollary}}{\end{corollary}}
\newenvironment{lem}{\begin{lemma}}{\end{lemma}}
\newenvironment{prop}{\begin{proposition}}{\end{proposition}}
\newenvironment{defn}{\begin{definition}}{\end{definition}}

\newcommand{\qed}{\hfill\BlackBox}

\floatstyle{ruled}
\newfloat{algorithm}{tbp}{loa}
\floatname{algorithm}{Algorithm}

\jmlrheading{}{2026}{1-\pageref{LastPage}}{}{}{}{Paul Geertsema and Helen Lu}

\ShortHeadings{AXIL: Exact Instance Attribution for Gradient Boosting}{Geertsema and Lu}
\firstpageno{1}
\begin{document}

\title{AXIL: Exact Instance Attribution for Gradient Boosting}

\author{\name Paul Geertsema \email paul.geertsema@vlerick.com \\
       \addr Vlerick Business School \\
       Reep 1, 9000 Ghent, Belgium
       \AND
       \name Helen Lu \email helen.lu@vlerick.com \\
       \addr Vlerick Business School \\
       Reep 1, 9000 Ghent, Belgium}

\editor{TBD}

\maketitle

\begin{abstract}%
We derive an exact, prediction-specific instance-attribution method for fitted gradient boosting machines (GBMs) trained with squared-error loss, with the learned tree structure held fixed. Each prediction can be written as a weighted sum of training targets, with coefficients determined only by the fitted tree structure and learning rate. These coefficients are exact instance attributions, or \emph{AXIL weights}. Our main algorithmic contribution is a matrix-free backward operator that computes one AXIL attribution vector in $\mathcal{O}(TN)$ time, or $S$ vectors in $\mathcal{O}(TNS)$, \emph{without} materialising the full $ N \times N $ matrix. This extends to out-of-sample predictions and makes exact instance attribution practical for large datasets. AXIL yields exact fixed-structure sensitivity by construction in target-perturbation tests, where competing GBM-specific attribution methods (BoostIn, TREX, and LeafInfluence) generally fail. In retraining-based faithfulness tests on 20 regression datasets, AXIL achieves the highest faithfulness score on 14 datasets and statistically ties for the best on 4 others, while also running substantially faster than the competing methods. We also show that the AXIL weight matrix is the globally constant special case of a target-response Jacobian that provides first-order instance attribution for any differentiable learner via implicit differentiation, placing the exact decomposition inside a broader framework. [198 words]
\end{abstract}

Code is available at \url{https://github.com/pgeertsema/AXIL_paper}.

\section{Introduction\label{sec:intro}}

When a machine learning model makes a prediction, the natural question is \emph{why}. Much of the Explainable AI (XAI) literature \citep{adadi2018peeking,gunning2019xai,xu2019explainable,arrieta2020explainable,linardatos2020explainable,molnar2020interpretable,kamath2021explainable} has focused on quantifying the significance of \emph{features} for particular predictions or models. This is understandable, as features are the driving force behind model predictions. Methods such as SHAP \citep{lundberg2017unified,lundberg2020local} and LIME \citep{ribeiro2016why} are widely used for this purpose. However, it is worthwhile reminding ourselves that the data matrix is two-dimensional, consisting of both features and instances. This observation motivates us to ask a complementary question: which \emph{training instances} drive a prediction?

We answer this question for gradient boosting machines (GBMs), the dominant supervised learning approach for tabular data \citep{borisov2021deep,grinsztajn2022why}. For a fitted squared-error GBM, we show that every prediction can be written as a linear combination of the training targets $ \boldsymbol{y} $:

\begin{align}
\widehat{y}_{i} & =\boldsymbol{k}_{i}\cdot\boldsymbol{y}=k_{i,1}y_{1}+k_{i,2}y_{2}+\ldots+k_{i,N}y_{N}\label{eq:AXIL weights formula}
\end{align}

The weight vector $\boldsymbol{k}_{i}$ is determined entirely by the fitted tree structure and learning rate. Since $\widehat{y}_{i}$ is linear in $\boldsymbol{y}$, each weight is the partial derivative of the prediction with respect to that training target:
\begin{equation}
k_{i,j} = \frac{\partial\,\widehat{y}_{i}}{\partial\,y_{j}}\bigg|_{\text{tree structure}}\label{eq:partial-deriv}
\end{equation}
AXIL therefore measures the fitted predictor's sensitivity to the training targets. It is not a leave-one-out, upweighting, or retraining effect. Concretely, $k_{i,j}$ is exactly how much prediction~$i$ would change if training target~$y_j$ were increased by one unit, \emph{with the tree structure held fixed}. A large positive $k_{i,j}$ means instance $j$ pulls prediction $i$ toward $y_j$; near-zero weights indicate negligible influence. We refer to these weights as \emph{AXIL weights} (Additive eXplanations with Instance Loadings).

For a single regression tree, linearity in $\boldsymbol{y}$ is immediate. The prediction for instance $i$ is the average of training targets in its leaf:
\begin{equation}
\boldsymbol{k}_{i}^{\text{TREE}}=\frac{1}{N_{i}}[1_{j\in\mathcal{L}_{i}}]_{j\in[1..N]}\label{eq:k-tree}
\end{equation}
where $N_i=|\mathcal{L}_{i}|$ and $\mathcal{L}_{i}$ is the set of instances in the same leaf as $i$. For a Random Forest \citep{breiman2001random} the weights are the average over $ T $ trees\footnote{These formulas assume each tree averages over the training rows present in that tree exactly once. For standard bootstrapped Random Forests, the same linear decomposition holds with multiplicity-adjusted leaf weights based on in-bag counts; we suppress that bookkeeping here.}:
\begin{equation}
\boldsymbol{k}_{i}^{\text{RF}}=\frac{1}{T}\sum_{t=1}^{T}\boldsymbol{k}_{i}^{\text{TREE},t}\label{eq:k-rf}
\end{equation}
The decomposition therefore covers regression trees and Random Forests directly (see also \citet{scornet2016random} who formalised this proximity structure as a kernel). For linear regression, assuming $\boldsymbol{X}$ has full column rank, $\widehat{\boldsymbol{y}}=\boldsymbol{H}\boldsymbol{y}$ where $\boldsymbol{H}=\boldsymbol{X}(\boldsymbol{X}^{T}\boldsymbol{X})^{-1}\boldsymbol{X}^{T}$ is the hat matrix \citep{hoaglin1978hat}, so the AXIL weights are the rows of $\boldsymbol{H}$. These are transparent linear cases. Our contribution is not merely to note that fitted GBMs are also linear in $\boldsymbol{y}$, but to turn that structure into exact, prediction-specific instance attributions and compute them tractably.

For GBMs the decomposition is more challenging, because each tree is trained on the residuals left by the previous trees, creating a chain of dependencies across the ensemble.\footnote{More generally, each GBM tree is trained on the \emph{gradient}
of the existing ensemble. For $L2$ loss the gradient equals the residuals
(see Section~10.10 of \citet{hastie2009elements}).} Theorem~\ref{thm:Any-GBM-regression} shows that, despite this complexity, linearity is preserved: the fitted GBM's predictions satisfy $\widehat{\boldsymbol{y}}=\boldsymbol{K}\boldsymbol{y}$ for a unique $N\times N$ AXIL weight matrix $\boldsymbol{K}$ whose $i$-th row is the AXIL weight vector $\boldsymbol{k}_{i}$, analogous to $\boldsymbol{H}$ in the linear case. The existence of $\boldsymbol{K}$ is the foundation; the substantive challenge is computational, namely to efficiently extract a single AXIL weight row vector.

The AXIL weight matrix $\boldsymbol{K}$ has $N^{2}$ entries. In principle one could compute the full matrix by propagating the boosting updates through all trees, but this quickly becomes infeasible at large $N$. A naive linear algebra implementation requires $\mathcal{O}(TN^{3})$ time. Even after exploiting the block-diagonal tree leaf structure, forming $\boldsymbol{K}$ still costs $\mathcal{O}(TN^{2})$ time, and at $N=1{,}000{,}000$ the matrix alone requires 8\,TB of memory. When the goal is to explain a single prediction, constructing the entire matrix is wasteful.

To address this, we develop a matrix-free backward pass through the fitted trees that computes the AXIL weight vector for any chosen prediction directly, without forming $\boldsymbol{K}$ (Theorem~\ref{thm:backward-operator}). The cost is $\mathcal{O}(TN)$ for one prediction and $\mathcal{O}(TNS)$ for $S$ predictions. Since $T$ (typically 100--500) and $S$ are much smaller than $N$, the per-prediction cost is effectively \emph{linear in the training set size}, making exact instance attribution practical for large datasets. For a fixed number of queried predictions, the procedure is asymptotically output-optimal: computing the requested weights is no slower in $\mathcal{O}(\cdot)$ terms than printing them out.

The paper also clarifies the boundary of this framework. We prove that the exact decomposition extends to classification trees and Random Forest classifiers with fitted structure held fixed (Corollary~\ref{cor:classification-trees-RF}). For GBM classifiers trained with log-loss on nondegenerate binary training sets with $N\geq 3$, the nonlinear base logit already rules out AXIL at initialisation and therefore at $T=1$, and for $T\geq 2$ only an exact cancellation by later trees could restore linearity (Theorem~\ref{thm:GBM-classification-impossible}). It is also provably impossible for a broad class of practically relevant neural networks formalised later in the paper (Proposition~\ref{prop:NN-impossible}). These results characterise where exact AXIL weight explanations are and are not possible.

However, the restriction to exact decomposition does not mean the underlying idea is limited to $L2$ GBMs. For any differentiable learner, the target-response Jacobian $\boldsymbol{J}=D_{\boldsymbol{y}}F(\boldsymbol{y})$ measures the first-order sensitivity of predictions to training targets, providing a local analogue of the AXIL weight matrix. For target-linear learners, $\boldsymbol{J}=\boldsymbol{K}$ and the attribution is exact and global; for general learners, $\boldsymbol{J}$ provides local, prediction-specific attribution computable via implicit differentiation under mild regularity conditions (Proposition~\ref{prop:target-response-jacobian}). This paper develops the exact case in full; the quality of the first-order approximation for general learners is an empirical question left for future work.

Empirically, we first verify the interpretation of AXIL weights in a target-perturbation experiment: AXIL exactly matches the true fixed-structure sensitivity by construction, whereas competing scores are materially less aligned with that quantity. We then evaluate faithfulness under retraining on 20 standard regression datasets. AXIL achieves the best faithfulness score on 14 datasets and is statistically tied for best on a further 4, with only 2 narrow losses (paired $t$-test, $\alpha=0.05$). Faithfulness is measured by a monotone data-removal protocol adapted from \citet{brophy2023adapting}: training instances are ranked by absolute attribution score, the top-ranked fraction is removed, the model is retrained, and the absolute change in that prediction is recorded as the area under the removal curve (AURC); higher AURC indicates better identification of truly influential instances.

Our contributions are therefore fivefold: (i)~an exact, prediction-specific instance-attribution framework for fitted squared-error GBM regression, including out-of-sample predictions, in which each prediction is a weighted sum of training targets, with AXIL weights assembling into a unique matrix $\boldsymbol{K}$ satisfying $\widehat{\boldsymbol{y}}=\boldsymbol{K}\boldsymbol{y}$ (Theorem~\ref{thm:Any-GBM-regression}); (ii)~a matrix-free backward operator computing any single AXIL weight vector in $\mathcal{O}(TN)$ time without forming $\boldsymbol{K}$, extending to $S$ predictions in $\mathcal{O}(TNS)$ and to out-of-sample predictions (Theorems~\ref{thm:backward-operator},~\ref{thm:oos-backward}); (iii)~boundary results characterising where this exact decomposition extends across the main model classes of interest, together with a simple criterion that unifies these cases (Proposition~\ref{prop:linear-leaf-update}, Table~\ref{tab:AXIL-applicability}); (iv)~experimental evidence that, unlike competing GBM attribution methods, AXIL matches exact target sensitivity in perturbation tests and leads on faithfulness score on 14 of 20 regression datasets, while being the fastest method in every comparison (Tables~\ref{tab:exactness}, \ref{tab:faithfulness}, \ref{tab:timings}); and~(v)~a formal connection showing that the AXIL weight matrix is the globally constant special case of the target-response Jacobian, an object that provides first-order instance attribution for any differentiable learner and is computable via implicit differentiation (Proposition~\ref{prop:target-response-jacobian}).

\section{Related work}

AXIL sits at the intersection of two literatures: methods that attribute model predictions to individual training instances, and results on when model predictions are linear in the training data. We review each in turn.

\subsection{Training-data attribution and influence}

\emph{Influence functions} \citep{koh2017understanding} estimate how upweighting a single training instance affects a test prediction, using implicit Hessian-vector products. The method is model-agnostic and widely used, but the resulting estimate is a first-order approximation and can be inaccurate when the perturbation is not infinitesimal.

\emph{TracIn} \citep{pruthi2020estimating} approximates training influence by summing gradient inner products along the training trajectory. For GBMs with $L2$ loss the per-step gradient is the residual, so TracIn reduces to a sum over trees of residual products between instances that share a leaf.

\emph{DataShapley} \citep{ghorbani2019data} and related data-valuation methods measure the marginal contribution of each training instance to overall model performance. These methods require many model retrainings and yield a single global score per instance, not prediction-specific weights.

\emph{Representer points} \citep{yeh2018representer} decompose a neural network's pre-activation predictions as a linear combination of training-point activations, with representer values as per-instance weights. The decomposition requires $L2$ regularisation on the output-layer weights and holds exactly for the pre-activation given convergence to a stationary point; it does not extend to the final post-nonlinearity output.

For gradient-boosted decision trees specifically, \citet{sharchilev2018finding} developed leaf-structured influence computations that exploit the tree leaf parameters to make influence calculations tractable. \citet{brophy2023adapting} bring together the main GBM-specific attribution baselines used in our experiments: \emph{BoostIn}, \emph{TREX}, and \emph{LeafInfluence}. BoostIn adapts TracIn by summing gradient inner products along the boosting trajectory; it measures gradient contributions rather than exact sensitivity to training targets and does not yield a linear decomposition of the prediction. TREX fits an $L2$-regularised kernel surrogate to the GBM's predictions and decomposes that surrogate via the representer theorem; like AXIL it produces a weighted sum of training instances, but the weights are approximate because the surrogate is not the GBM. LeafInfluence adapts influence-style calculations to the leaf-parameter structure of GBMs. These three methods form the natural experimental comparison set for AXIL.

\subsection{Linearity in the training data}

In statistical learning theory, \citet{buhlmann2003boosting} proved that $L2$ boosting with linear base learners produces a \emph{linear smoother}, meaning predictions satisfy $\widehat{\boldsymbol{y}}=\boldsymbol{S}\boldsymbol{y}$ for a fixed smoothing matrix $\boldsymbol{S}$, a property used to characterise model degrees of freedom. A regression tree with fixed structure is itself a linear operator, since it maps targets to leaf averages, so the same existence idea carries over to fitted squared-error GBMs. Our contribution is not merely the existence of such a matrix. Rather, we identify its rows as exact, prediction-specific instance attributions, derive an explicit recursion for it, show how to extract any single row in $\mathcal{O}(TN)$ time via a backward operator, extend the construction to out-of-sample predictions, and characterise where the exact decomposition does and does not extend.

A related line of work asks whether model predictions are linear in the training data more broadly. \citet{ilyas2022datamodels} introduced \emph{datamodels}, showing empirically that deep neural-network predictions are approximately linear in training-set membership indicators, and that this approximation can be highly predictive even for complex models. \citet{park2023trak} developed TRAK, which makes this approximation computationally tractable at scale using random projections. These results are empirical approximations for neural networks. By contrast, for fitted squared-error GBMs we prove an \emph{exact} linear relationship in the training targets, and the boundary results of Section~\ref{sec:boundary-results} show where this exactness does and does not extend. More broadly, implicit differentiation of the training objective provides the derivative of model predictions with respect to training targets for any twice-differentiable parametric learner \citep{lorraine2020optimizing,franceschi2018bilevel}. This target-response Jacobian is a general object of which AXIL's weight matrix $\boldsymbol{K}$ is the globally constant special case (Proposition~\ref{prop:target-response-jacobian}).

Overall, the closest prior work to AXIL consists of GBM-specific training-data attribution methods and the literature on linear dependence on the training data. Relative to that work, AXIL provides \emph{exact}, \emph{prediction-specific} instance weights for squared-error GBM regression together with a scalable matrix-free algorithm for computing them. This positioning also motivates our benchmark choice in Section~\ref{sec:experiments}: we compare primarily against BoostIn, TREX, and LeafInfluence, because they are the closest GBM-specific, prediction-specific attribution methods, whereas methods such as DataShapley are global rather than prediction-specific and methods such as TRAK target approximate training-data attribution in neural networks rather than tree boosting.

%======================================================================
\section{The AXIL Decomposition}\label{sec:decomposition}
%======================================================================

In this section we study the map from training targets $\boldsymbol{y}$ to predictions induced by a \emph{fitted} GBM. Unless stated otherwise, the fitted ensemble structure is held fixed: this means the learned tree topology, leaf memberships, and, under row subsampling, the contributing sets that determine leaf values. Under this standing convention, every GBM regression prediction is a linear combination of the training targets $\boldsymbol{y}$. The key observation is that the remaining operations in the GBM update (leaf averaging, addition, and scaling by the learning rate $\lambda$) are linear in $\boldsymbol{y}$. This allows us to track an explicit AXIL weight matrix $\boldsymbol{K}$ through the step-by-step ensemble update. The result formalises the linear smoother property of $L2$ boosting \citep{buhlmann2003boosting}, and the recursive structure thereby exposed also underpins the efficient algorithms of Section~\ref{sec:computing}. A complete worked example for a minimal case ($N=4$, $T=2$) is given in Appendix~\ref{subsec:worked-example}; it covers the AXIL recursion and the key algorithms developed in this section and Section~\ref{sec:computing}.

\subsection{The leaf-averaging operator}

Each tree in a GBM ensemble partitions the training instances into leaves. The \emph{leaf-averaging operator} $\boldsymbol{W}_{t}$ for tree $t$ replaces each entry of a vector with the mean over entries in the same leaf. Let $\mathcal{L}_{i}^{(t)}$ denote the set of training instances assigned to the same leaf as instance $i$ in tree $t$. In the \emph{full-batch} setting (no row subsampling; used throughout our experiments):
\begin{equation}
(\boldsymbol{W}_{t}\boldsymbol{v})_{i}=\frac{1}{|\mathcal{L}_{i}^{(t)}|}\sum_{j\in\mathcal{L}_{i}^{(t)}}v_{j}\label{eq:W-def}
\end{equation}
For a tree with $N=5$ instances and two leaves $\mathcal{L}_{1}=\{1,2\}$, $\mathcal{L}_{2}=\{3,4,5\}$, the matrix representation is:
\begin{equation}\label{eq:W-example}
\boldsymbol{W} = \begin{pmatrix}\tfrac{1}{2}&\tfrac{1}{2}&0&0&0\\[2pt]\tfrac{1}{2}&\tfrac{1}{2}&0&0&0\\[2pt]0&0&\tfrac{1}{3}&\tfrac{1}{3}&\tfrac{1}{3}\\[2pt]0&0&\tfrac{1}{3}&\tfrac{1}{3}&\tfrac{1}{3}\\[2pt]0&0&\tfrac{1}{3}&\tfrac{1}{3}&\tfrac{1}{3}\end{pmatrix}
\end{equation}
Instances in the same leaf average together; cross-leaf entries are zero. Each block has identical rows (every instance in a leaf gets the same average) and each row sums to one (the leaf mean is a convex combination of its entries).\footnote{When row subsampling is used, only a random subset $\mathcal{S}_{t}\subseteq\{1,\ldots,N\}$ of training instances contributes to each tree's leaf values. In this case the leaf-averaging operator averages over the \emph{contributing set} $\mathcal{C}_{i}^{(t)}=\mathcal{S}_{t}\cap\mathcal{L}_{i}^{(t)}$ rather than the full leaf $\mathcal{L}_{i}^{(t)}$, and $\boldsymbol{W}_{t}$ is no longer symmetric in general. The same decomposition and $\mathcal{O}(TN)$ backward recursion still apply because each tree still induces a fixed linear operator, but the backward pass must use $\boldsymbol{W}_{t}^{T}$ and column sums of $\boldsymbol{K}$ need not equal one. LightGBM does not expose the subsample indices, so our implementation and experiments use the full-batch setting throughout.}

\begin{lem}[Leaf-averaging operator]\label{lem:block-diagonal}
For any tree $t$, the leaf-averaging operator $\boldsymbol{W}_{t}$ is a fixed linear map on $\mathbb{R}^{N}$, determined solely by tree $t$'s fitted leaf assignment and, under row subsampling, the associated contributing sets $\{\mathcal{C}_{i}^{(t)}\}_{i=1}^{N}$. Although $\boldsymbol{W}_{t}$ can be represented as an $N\times N$ matrix, it need never be formed explicitly: both $\boldsymbol{W}_{t}$ and $\boldsymbol{W}_{t}^{T}$ can be applied to an $N$-vector in $\mathcal{O}(N)$ time. In the row-subsampled setting, the transposed application uses the contributing sets.
\end{lem}
\begin{proof}
See Appendix~\ref{subsec:W-structural-props}.
\end{proof}

\subsection{The AXIL recursion\label{sec:recursion}}

Let $\widehat{\boldsymbol{y}}^{(t)}$ denote the ensemble's predictions after $t$ trees, and let $\boldsymbol{r}^{(t-1)}=\boldsymbol{y}-\widehat{\boldsymbol{y}}^{(t-1)}$ be the current residuals. Tree $t$ replaces each residual with the mean residual in its leaf, then the ensemble is updated:
\begin{align}
\widehat{\boldsymbol{y}}^{(0)} &= \overline{y}\cdot\boldsymbol{1} \label{eq:yhat0}\\[4pt]
\widehat{\boldsymbol{y}}^{(t)} &= \widehat{\boldsymbol{y}}^{(t-1)} + \lambda\,\boldsymbol{W}_{t}\underbrace{\bigl(\boldsymbol{y}-\widehat{\boldsymbol{y}}^{(t-1)}\bigr)}_{\boldsymbol{r}^{(t-1)}},\quad t=1,\ldots,T \label{eq:yhat-vec}
\end{align}

Since each $\boldsymbol{W}_{t}$ is a fixed linear operator (Lemma~\ref{lem:block-diagonal}), the GBM mechanics~(\ref{eq:yhat0})--(\ref{eq:yhat-vec}) directly determine a matrix recursion for the map from training targets to predictions. Rewriting the initialisation~(\ref{eq:yhat0}):
\begin{equation}
\widehat{\boldsymbol{y}}^{(0)}=\overline{y}\cdot\boldsymbol{1}=\tfrac{1}{N}(\boldsymbol{1}^{T}\boldsymbol{y})\,\boldsymbol{1}=\underbrace{\tfrac{1}{N}\boldsymbol{1}\boldsymbol{1}^{T}}_{\boldsymbol{K}_{0}}\boldsymbol{y}\label{eq:K0}
\end{equation}
For subsequent steps, substituting $\widehat{\boldsymbol{y}}^{(t-1)}=\boldsymbol{K}_{t-1}\boldsymbol{y}$ into the update rule~(\ref{eq:yhat-vec}):
\begin{align}
\widehat{\boldsymbol{y}}^{(t)}
&=\boldsymbol{K}_{t-1}\boldsymbol{y}+\lambda\,\boldsymbol{W}_{t}\bigl(\boldsymbol{y}-\boldsymbol{K}_{t-1}\boldsymbol{y}\bigr)
  &&\text{[substitute into~(\ref{eq:yhat-vec})]}\nonumber\\
&=\bigl[\boldsymbol{K}_{t-1}+\lambda\,\boldsymbol{W}_{t}(\boldsymbol{I}-\boldsymbol{K}_{t-1})\bigr]\boldsymbol{y}
  &&\text{[factor out $\boldsymbol{y}$]}\nonumber
\end{align}
Identifying the bracketed matrix as $\boldsymbol{K}_{t}$ gives the AXIL matrix recursion:
\begin{equation}
\boldsymbol{K}_{t} = \underbrace{\boldsymbol{K}_{t-1}}_{\substack{\text{ensemble}\\\text{weights so far}}} +\;\underbrace{\lambda}_{\substack{\text{learning}\\\text{rate}}}\;\underbrace{\boldsymbol{W}_{t}}_{\substack{\text{leaf avg,}\\\text{tree }t}}\underbrace{\left(\boldsymbol{I}-\boldsymbol{K}_{t-1}\right)}_{\substack{\text{unexplained}\\\text{weights}}},\quad t=1,\ldots,T\label{eq:Gt-recursion}
\end{equation}
Write $\boldsymbol{K}:=\boldsymbol{K}_{T}$ for the terminal matrix.

\begin{thm}[The AXIL decomposition]\label{thm:Any-GBM-regression}
For any fitted GBM regressor with squared-error loss, with the fitted tree structures $\boldsymbol{W}_1,\ldots,\boldsymbol{W}_T$ held fixed, $\widehat{\boldsymbol{y}}=\boldsymbol{K}\boldsymbol{y}$ for every $\boldsymbol{y}\in\mathbb{R}^{N}$, and $\boldsymbol{K}$ is the unique matrix with this property. The matrix depends only on the fitted tree structure and the learning rate, and its $i$-th row is the AXIL weight vector $\boldsymbol{k}_{i}$. Moreover, with the fitted tree structures held fixed, any out-of-sample prediction is also a linear function of the training targets $\boldsymbol{y}$.
\end{thm}
\begin{proof}
\emph{Existence.} By induction on $t$. The base case holds by~(\ref{eq:K0}): $\widehat{\boldsymbol{y}}^{(0)}=\boldsymbol{K}_{0}\boldsymbol{y}$. For the inductive step, assume $\widehat{\boldsymbol{y}}^{(t-1)}=\boldsymbol{K}_{t-1}\boldsymbol{y}$; then the derivation of~(\ref{eq:Gt-recursion}) above gives $\widehat{\boldsymbol{y}}^{(t)}=\boldsymbol{K}_{t}\boldsymbol{y}$.

\emph{Uniqueness.} If $\boldsymbol{M}$ also satisfies $\widehat{\boldsymbol{y}}=\boldsymbol{M}\boldsymbol{y}$ for every $\boldsymbol{y}$, then setting $\boldsymbol{y}=\boldsymbol{e}_j$, where $\boldsymbol{e}_j$ denotes the $j$th standard basis vector in $\mathbb{R}^N$, gives column~$j$ of $\boldsymbol{M}$ equal to column~$j$ of $\boldsymbol{K}$, for each $j$. Hence $\boldsymbol{M}=\boldsymbol{K}$.

\emph{Out-of-sample extension.} The same argument extends to any instance $\boldsymbol{x}$, not necessarily in the training set. For such an $\boldsymbol{x}$, tree $t$ predicts the mean of the residuals $\boldsymbol{r}^{(t-1)}$ over the leaf containing $\boldsymbol{x}$. This is a fixed linear combination of entries of $\boldsymbol{r}^{(t-1)}=(\boldsymbol{I}-\boldsymbol{K}_{t-1})\boldsymbol{y}$, hence linear in $\boldsymbol{y}$. The full prediction, $\overline{y}$ plus $\lambda$ times the sum of tree predictions, is therefore linear in $\boldsymbol{y}$.
\end{proof}

The proof uses only that each $\boldsymbol{W}_{t}$ is a fixed linear operator; the decomposition therefore extends to any GBM variant with linear leaf updates, including $L2$ leaf regularisation and the row-subsampled setting described in Section~3.1. It does not extend to variants with nonlinear leaf transforms such as $L1$ regularisation (Section~\ref{sec:boundary-results}).\footnote{If the implementation applies $L2$ regularisation to leaf values, replacing the leaf mean with $\sum_{j\in\mathcal{L}} r_j / (n_l + \lambda_{\mathrm{reg}})$, the leaf operation remains linear in $\boldsymbol{y}$ and the decomposition $\widehat{\boldsymbol{y}}=\boldsymbol{K}\boldsymbol{y}$ still holds (with a different $\boldsymbol{K}$), but $\boldsymbol{W}_t$ is no longer idempotent or row-stochastic (Appendix~\ref{subsec:W-structural-props}). $L1$ leaf regularisation introduces nonlinear soft-thresholding and breaks the decomposition. Our experiments use LightGBM's defaults ($\lambda_{L1}=\lambda_{L2}=0$).}

We now derive two rearrangements of the recursion~(\ref{eq:Gt-recursion}) which will be useful for the efficient algorithms of Section~\ref{sec:computing}.

\emph{First}, we obtain a factored form that underpins the backward operator (Theorem~\ref{thm:backward-operator}). Expanding the product $\lambda\boldsymbol{W}_{t}(\boldsymbol{I}-\boldsymbol{K}_{t-1})$ in~(\ref{eq:Gt-recursion}) and collecting terms in $\boldsymbol{K}_{t-1}$:
\begin{equation}
\boldsymbol{K}_{t} = \boldsymbol{K}_{t-1}+\lambda\boldsymbol{W}_{t}-\lambda\boldsymbol{W}_{t}\boldsymbol{K}_{t-1} = (\boldsymbol{I}-\lambda\boldsymbol{W}_{t})\,\boldsymbol{K}_{t-1}+\lambda\boldsymbol{W}_{t},\quad t=1,\ldots,T\label{eq:Kt-factored-form}
\end{equation}
The intuitive form~(\ref{eq:Gt-recursion}) shows each tree absorbing unexplained signal; the factored form~(\ref{eq:Kt-factored-form}) isolates $\boldsymbol{K}_{t-1}$ as a right factor, which is the form used by the backward operator (Section~\ref{subsec:backward-operator}). 

\emph{Second} we construct a residual recursion used in the out-of-sample extension (Theorem~\ref{thm:oos-backward}). Subtracting~(\ref{eq:Kt-factored-form}) from $\boldsymbol{I}$ gives $\boldsymbol{I}-\boldsymbol{K}_{t}=(\boldsymbol{I}-\lambda\boldsymbol{W}_{t})(\boldsymbol{I}-\boldsymbol{K}_{t-1})$: each tree shrinks the residual vector multiplicatively by the factor $(\boldsymbol{I}-\lambda\boldsymbol{W}_{t})$. Writing $\boldsymbol{r}^{(t)}:=(\boldsymbol{I}-\boldsymbol{K}_{t})\boldsymbol{y}$ for the residuals (with $\boldsymbol{r}^{(0)}=\boldsymbol{y}-\overline{y}\boldsymbol{1}$), right-multiplying by $\boldsymbol{y}$ gives:
\begin{equation}
\boldsymbol{r}^{(t)}=(\boldsymbol{I}-\lambda\boldsymbol{W}_{t})\boldsymbol{r}^{(t-1)},\quad t=1,\ldots,T\label{eq:r-recursion}
\end{equation}
The full residual after $T$ trees is therefore $\prod_{t=1}^{T}(\boldsymbol{I}-\lambda\boldsymbol{W}_{t})\cdot(\boldsymbol{I}-\boldsymbol{K}_{0})$ applied to $\boldsymbol{y}$.

\paragraph{Key properties of $\boldsymbol{K}$}
Every row of $\boldsymbol{K}$ sums to one, so shifting all targets by a constant shifts every prediction by the same amount. In the full-batch setting every column also sums to one (\emph{double stochasticity}), meaning every training instance exerts equal aggregate influence across all predictions (proof in Appendix~\ref{subsec:conservation}). Individual weights $k_{i,j}$ can be negative: a single tree averages targets in each leaf (non-negative weights), but in a GBM each tree averages \emph{residuals}, and tracing the weights back to $\boldsymbol{y}$ through the recursion introduces subtractive terms.

\begin{figure}[h!t]
\centering
\includegraphics[width=\textwidth]{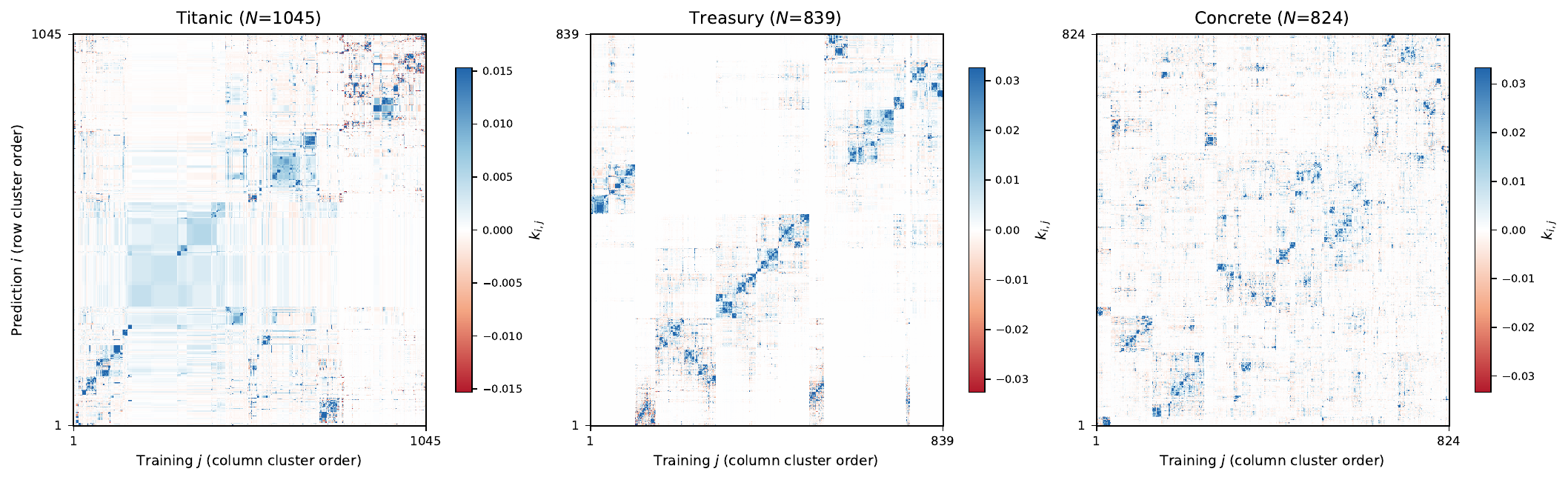}
\caption{AXIL weight matrix $\boldsymbol{K}$ for three datasets, with rows and columns in hierarchical cluster order. Diverging colours: $k_{i,j}>0$ (blue), $k_{i,j}<0$ (red). The colour scale uses symmetric limits $\pm$ the 99th percentile of $|k_{i,j}|$ so bulk structure is visible; the largest 1\% of $|k_{i,j}|$ saturate at the ends of the bar.\label{fig:K-heatmaps}}
\end{figure}

In practice one often needs weights for only $S$ predictions, which may include out-of-sample instances (Section~\ref{subsec:oos}). The relevant object is then an $S\times N$ matrix whose rows are the AXIL weight vectors $\boldsymbol{k}_{i_1},\ldots,\boldsymbol{k}_{i_S}$, each of length $N$. When $S=N$ the matrix is the full square $\boldsymbol{K}$. Figure~\ref{fig:K-heatmaps} visualises $\boldsymbol{K}$ for three representative datasets, with rows and columns reordered by hierarchical clustering. The block-diagonal structure reflects groups of instances that share leaves across many trees and therefore exert strong mutual influence.

%======================================================================
\section{Efficient Computation}\label{sec:computing}
%======================================================================

Working under the fixed-structure viewpoint established in Section~\ref{sec:decomposition}, the AXIL recursion~(\ref{eq:Gt-recursion}) defines the AXIL weight matrix $\boldsymbol{K}$, but forming $\boldsymbol{K}$ explicitly is expensive, and obtaining weights for a single prediction requires an entirely different approach. This section develops the backward operator, which computes exact AXIL weights for any single prediction in $\mathcal{O}(TN)$ time without materialising $\boldsymbol{K}$, and extends it to out-of-sample predictions.

\subsection{The computational challenge\label{subsec:comp-challenge}}

Because $\boldsymbol{W}_{t}$ is block-diagonal (Appendix~\ref{subsec:W-structural-props}), applying it to a dense $N\times N$ matrix costs $\mathcal{O}(N^{2})$ rather than $\mathcal{O}(N^{3})$, so the full AXIL weight matrix $\boldsymbol{K}$ can be computed via the recursion~(\ref{eq:Gt-recursion}) in $\mathcal{O}(TN^{2})$ time, down from a na\"ive $\mathcal{O}(TN^{3})$. But $\boldsymbol{K}$ has $N^{2}$ entries; at $N=1{,}000{,}000$ it requires 8\,TB of memory, which is a stretch on most hardware.

Applying the recursion to a single \emph{vector} $\boldsymbol{v}$ instead of the full identity gives $\boldsymbol{K}\boldsymbol{v}$ in $\mathcal{O}(TN)$, since each step replaces a matrix multiplication with a leaf-averaging operation on an $N$-vector (Appendix~\ref{subsec:forward-operator-appendix}). We call this the \emph{forward operator}. Setting $\boldsymbol{v}=\boldsymbol{y}$ reproduces the GBM's predictions; setting $\boldsymbol{v}=\boldsymbol{e}_{j}$ yields \emph{column~$j$} of $\boldsymbol{K}$, the vector of influences that training target $y_{j}$ exerts on every prediction.

For explanation of prediction $i$, however, we need \emph{row~$i$} of $\boldsymbol{K}$, not a column. Extracting row~$i$ via the forward operator would require $N$ calls (one per column), costing $\mathcal{O}(TN^{2})$ overall. Nor can we read rows from columns, because $\boldsymbol{K}$ is not symmetric in general: $\boldsymbol{K}_{1}$ is symmetric for a single tree, but for $T\geq 2$ symmetry fails whenever $\boldsymbol{W}_{1}\boldsymbol{W}_{2}\neq\boldsymbol{W}_{2}\boldsymbol{W}_{1}$, which holds for generic leaf partitions. Since row~$i$ of $\boldsymbol{K}$ equals $\boldsymbol{K}^{T}\boldsymbol{e}_{i}$, we need an efficient operator for $\boldsymbol{K}^{T}$. The formalisation of this operator is our main theoretical contribution.

\subsection{The backward operator\label{subsec:backward-operator}}

\begin{thm}[Backward operator]\label{thm:backward-operator}
For any $\boldsymbol{u}\in\mathbb{R}^{N}$, define the backward recursion:
\begin{align}
\boldsymbol{h}_{T} &= \boldsymbol{u}\label{eq:backward-init}\\
\boldsymbol{h}_{t-1} &= \boldsymbol{h}_{t}-\lambda\,\boldsymbol{W}_{t}^{T}\boldsymbol{h}_{t},\quad t=T,\ldots,1\label{eq:backward-step}
\end{align}
Then
\begin{equation}
\boldsymbol{K}^{T}\boldsymbol{u}=\overline{h_{0}}\cdot\boldsymbol{1}+\sum_{t=1}^{T}\lambda\,\boldsymbol{W}_{t}^{T}\boldsymbol{h}_{t}\label{eq:backward-result}
\end{equation}
where $\overline{h_{0}}=\frac{1}{N}\boldsymbol{1}^{T}\boldsymbol{h}_{0}$. In the full-batch setting $\boldsymbol{W}_{t}^{T}=\boldsymbol{W}_{t}$, so each step is a standard leaf-averaging operation; more generally, when the contributing sets are available, Lemma~\ref{lem:block-diagonal} gives $\mathcal{O}(N)$ cost per transposed application. The total cost is $\mathcal{O}(TN)$.
\end{thm}
\begin{proof}
Transposing the factored recursion~(\ref{eq:Kt-factored-form}) and right-multiplying by $\boldsymbol{h}_{t}$:
\begin{align}
\boldsymbol{K}_{t}^{T}\boldsymbol{h}_{t}
&=\bigl[\boldsymbol{K}_{t-1}^{T}(\boldsymbol{I}-\lambda\boldsymbol{W}_{t}^{T})
  +\lambda\boldsymbol{W}_{t}^{T}\bigr]\boldsymbol{h}_{t}
  &&\text{[transpose~(\ref{eq:Kt-factored-form}), multiply by $\boldsymbol{h}_{t}$]}\nonumber\\
&=\boldsymbol{K}_{t-1}^{T}\underbrace{(\boldsymbol{I}-\lambda\boldsymbol{W}_{t}^{T})\boldsymbol{h}_{t}}_{\boldsymbol{h}_{t-1}\text{ by~(\ref{eq:backward-step})}}
  +\lambda\boldsymbol{W}_{t}^{T}\boldsymbol{h}_{t}
  &&\text{[distribute]}\nonumber\\
&=\boldsymbol{K}_{t-1}^{T}\boldsymbol{h}_{t-1}
  +\lambda\boldsymbol{W}_{t}^{T}\boldsymbol{h}_{t}\label{eq:Kh-step}
\end{align}
Rearranging:
\begin{equation}
\boldsymbol{K}_{t}^{T}\boldsymbol{h}_{t}-\boldsymbol{K}_{t-1}^{T}\boldsymbol{h}_{t-1}=\lambda\,\boldsymbol{W}_{t}^{T}\boldsymbol{h}_{t}\label{eq:Kh-telescope-step}
\end{equation}
Summing over $t=1,\ldots,T$, the left side telescopes:
\begin{equation}
\boldsymbol{K}_{T}^{T}\boldsymbol{h}_{T}-\boldsymbol{K}_{0}^{T}\boldsymbol{h}_{0}=\sum_{t=1}^{T}\lambda\,\boldsymbol{W}_{t}^{T}\boldsymbol{h}_{t}\label{eq:Kh-telescope}
\end{equation}
Substituting $\boldsymbol{h}_{T}=\boldsymbol{u}$ (by~(\ref{eq:backward-init})) and noting that $\boldsymbol{K}_{0}=\frac{1}{N}\boldsymbol{1}\boldsymbol{1}^{T}$ is symmetric, so $\boldsymbol{K}_{0}^{T}\boldsymbol{h}_{0}=\frac{1}{N}\boldsymbol{1}(\boldsymbol{1}^{T}\boldsymbol{h}_{0})=\overline{h_{0}}\cdot\boldsymbol{1}$:
\begin{equation}
\boldsymbol{K}_{T}^{T}\boldsymbol{u}=\overline{h_{0}}\cdot\boldsymbol{1}+\sum_{t=1}^{T}\lambda\,\boldsymbol{W}_{t}^{T}\boldsymbol{h}_{t}
\end{equation}
which is~(\ref{eq:backward-result}) as required. Each step applies $\boldsymbol{W}_{t}^{T}$ to an $N$-vector, costing $\mathcal{O}(N)$ (Lemma~\ref{lem:block-diagonal}), giving $\mathcal{O}(TN)$ total.
\end{proof}

The AXIL weight vector for prediction~$i$ is row~$i$ of $\boldsymbol{K}$, which equals $\boldsymbol{K}^{T}\boldsymbol{e}_{i}$. Setting $\boldsymbol{u}=\boldsymbol{e}_{i}$ in Theorem~\ref{thm:backward-operator} therefore computes $\boldsymbol{k}_{i}$ in $\mathcal{O}(TN)$ time. For $S$ predictions, running the backward operator independently on $S$ basis vectors costs $\mathcal{O}(TNS)$; when $S=N$ this produces the full $\boldsymbol{K}$ in $\mathcal{O}(TN^{2})$.

\subsection{Out-of-sample predictions\label{subsec:oos}}

In practice, most predictions of interest are out-of-sample. Let $f_{T}(\boldsymbol{x}_{\mathrm{new}})$ denote the fitted GBM's prediction for a new instance $\boldsymbol{x}_{\mathrm{new}}$ after $T$ trees. By Theorem~\ref{thm:Any-GBM-regression}, under this same fixed fitted ensemble, every such prediction is linear in $\boldsymbol{y}$, so there exists an AXIL weight vector $\boldsymbol{k}_{\mathrm{new}}$ with $f_{T}(\boldsymbol{x}_{\mathrm{new}})=\boldsymbol{k}_{\mathrm{new}}^{T}\boldsymbol{y}$. To compute $\boldsymbol{k}_{\mathrm{new}}$ efficiently, we write the prediction explicitly. The GBM initialises at $\overline{y}$ and at each step adds $\lambda$ times the leaf mean of the current residuals; for a new instance $\boldsymbol{x}_{\mathrm{new}}$, the leaf mean of the residuals $\boldsymbol{r}^{(t-1)}$ in tree $t$ is $\boldsymbol{c}_{t}^{T}\boldsymbol{r}^{(t-1)}$ by definition of $\boldsymbol{c}_{t}$. Summing over trees gives:
\begin{equation}
f_{T}(\boldsymbol{x}_{\mathrm{new}}) = \overline{y} + \sum_{t=1}^{T}\lambda\, \boldsymbol{c}_{t}^{T}\boldsymbol{r}^{(t-1)}\label{eq:oos-prediction}
\end{equation}
where $\boldsymbol{c}_{t}\in\mathbb{R}^{N}$ is the \emph{cross-leaf vector} for tree $t$: in the full-batch setting, $c_{t,j}=1/n_{l}$ if training instance $j$ shares a leaf with $\boldsymbol{x}_{\mathrm{new}}$ in tree $t$ (with $n_{l}$ training instances in that leaf), and $0$ otherwise. Under row subsampling, the same definition uses the contributing set $\mathcal{C}_{l}^{(t)}$ for the relevant leaf: $c_{t,j}=1/|\mathcal{C}_{l}^{(t)}|$ for $j\in\mathcal{C}_{l}^{(t)}$ and $0$ otherwise. Each $\boldsymbol{c}_{t}$ depends only on $\boldsymbol{x}_{\mathrm{new}}$ and the learned splits, so it is fixed once the tree structure is fixed. The residuals $\boldsymbol{r}^{(t-1)}=(\boldsymbol{I}-\boldsymbol{K}_{t-1})\boldsymbol{y}$ are the in-sample residuals from Theorem~\ref{thm:Any-GBM-regression}. The following theorem gives an $\mathcal{O}(TN)$ backward recursion for $\boldsymbol{k}_{\mathrm{new}}$, structurally parallel to Theorem~\ref{thm:backward-operator}: the same $(\boldsymbol{I}-\lambda\boldsymbol{W}_{t}^{T})$ factor appears at each step, with an additional source term $\lambda\boldsymbol{c}_{t}$ that injects the cross-leaf information for each tree.

\begin{thm}[Out-of-sample backward operator]\label{thm:oos-backward}
For a new instance $\boldsymbol{x}_{\mathrm{new}}$ with cross-leaf vectors $\boldsymbol{c}_{1},\ldots,\boldsymbol{c}_{T}$, define the backward recursion:
\begin{align}
\boldsymbol{p}_{T+1} &= \boldsymbol{0}\label{eq:oos-backward-init}\\
\boldsymbol{p}_{t} &= \lambda\boldsymbol{c}_{t}+(\boldsymbol{I}-\lambda\boldsymbol{W}_{t}^{T})\boldsymbol{p}_{t+1},\quad t=T,\ldots,1\label{eq:oos-backward-step}
\end{align}
Then
\begin{equation}
\boldsymbol{k}_{\mathrm{new}}=\boldsymbol{p}_{1}+\frac{1-\boldsymbol{1}^{T}\boldsymbol{p}_{1}}{N}\boldsymbol{1}\label{eq:oos-backward-result}
\end{equation}
In the full-batch setting $\boldsymbol{W}_{t}^{T}=\boldsymbol{W}_{t}$, so each step is a leaf-averaging operation plus a vector addition; more generally, when the contributing sets are available, Lemma~\ref{lem:block-diagonal} gives $\mathcal{O}(N)$ cost for the transposed application. The total cost is $\mathcal{O}(TN)$.
\end{thm}
\begin{proof}
Substituting~(\ref{eq:oos-backward-step}) into the inner product $\boldsymbol{p}_{t}^{T}\boldsymbol{r}^{(t-1)}$, where $\boldsymbol{r}^{(t)}$ is the residual recursion~(\ref{eq:r-recursion}):
\begin{align}
\boldsymbol{p}_{t}^{T}\boldsymbol{r}^{(t-1)}
&=\bigl[\lambda\boldsymbol{c}_{t}+(\boldsymbol{I}-\lambda\boldsymbol{W}_{t}^{T})\boldsymbol{p}_{t+1}\bigr]^{T}\boldsymbol{r}^{(t-1)}
  &&\text{[substitute~(\ref{eq:oos-backward-step})]}\nonumber\\
&=\lambda\boldsymbol{c}_{t}^{T}\boldsymbol{r}^{(t-1)}
  +\boldsymbol{p}_{t+1}^{T}\underbrace{(\boldsymbol{I}-\lambda\boldsymbol{W}_{t})\boldsymbol{r}^{(t-1)}}_{\boldsymbol{r}^{(t)}\text{ by~(\ref{eq:r-recursion})}}
  &&\text{[distribute; transpose]}\label{eq:pr-step}
\end{align}
So $\boldsymbol{p}_{t}^{T}\boldsymbol{r}^{(t-1)}-\boldsymbol{p}_{t+1}^{T}\boldsymbol{r}^{(t)}=\lambda\boldsymbol{c}_{t}^{T}\boldsymbol{r}^{(t-1)}$. Summing over $t=1,\ldots,T$ telescopes the left side; with $\boldsymbol{p}_{T+1}=\boldsymbol{0}$ from~(\ref{eq:oos-backward-init}), the right side equals $f_{T}(\boldsymbol{x}_{\mathrm{new}})-\overline{y}$ by~(\ref{eq:oos-prediction}):
\begin{equation}
\boldsymbol{p}_{1}^{T}\boldsymbol{r}^{(0)}=f_{T}(\boldsymbol{x}_{\mathrm{new}})-\overline{y}\label{eq:pr-telescoped}
\end{equation}
Substituting $\boldsymbol{r}^{(0)}=\boldsymbol{y}-\overline{y}\boldsymbol{1}$ and $\overline{y}=\frac{1}{N}\boldsymbol{1}^{T}\boldsymbol{y}$:
\begin{equation}
f_{T}(\boldsymbol{x}_{\mathrm{new}})=\overline{y}+\boldsymbol{p}_{1}^{T}(\boldsymbol{y}-\overline{y}\boldsymbol{1})=\Bigl[\boldsymbol{p}_{1}+\frac{1-\boldsymbol{1}^{T}\boldsymbol{p}_{1}}{N}\boldsymbol{1}\Bigr]^{T}\boldsymbol{y}
\end{equation}
Since $f_{T}(\boldsymbol{x}_{\mathrm{new}})=\boldsymbol{k}_{\mathrm{new}}^{T}\boldsymbol{y}$ for all $\boldsymbol{y}$, reading off the weight vector gives~(\ref{eq:oos-backward-result}) as required. Each step applies $\boldsymbol{W}_{t}^{T}$ to an $N$-vector and adds $\lambda\boldsymbol{c}_{t}$, both $\mathcal{O}(N)$, giving $\mathcal{O}(TN)$ total.
\end{proof}

Both backward operators depend only on the leaf membership vectors $\boldsymbol{\ell}_{1},\ldots,\boldsymbol{\ell}_{T}$ (recording which leaf each training instance was assigned to in each tree) and the learning rate~$\lambda$. No feature values, gradients, or Hessians are needed after the model is fit. For out-of-sample predictions, the queried instance's leaf assignments are additionally required. This distinguishes AXIL from influence-function methods, which require access to gradients or Hessians at explanation time.

%======================================================================
\section{Algorithms\label{sec:algorithms}}
%======================================================================

The forward and backward operators both rely on a single subroutine, \textsc{LeafAverage}, which replaces each entry of a vector with the mean over its leaf. This section presents the four AXIL algorithms.

\textbf{Scope.} The algorithms below implement the full-batch setting ($\mathcal{C}_{i}^{(t)}=\mathcal{L}_{i}^{(t)}$ for all $i,t$), in which $\boldsymbol{W}_{t}$ is symmetric and $\boldsymbol{W}_{t}^{T}\boldsymbol{v}=\boldsymbol{W}_{t}\boldsymbol{v}$, so \textsc{LeafAverage} correctly implements both $\boldsymbol{W}_{t}\boldsymbol{v}$ and $\boldsymbol{W}_{t}^{T}\boldsymbol{v}$. Under row subsampling $\boldsymbol{W}_{t}^{T}\neq\boldsymbol{W}_{t}$, so a correct implementation would additionally require the per-tree contributing-set indicators $\mathcal{C}_{i}^{(t)}$, which LightGBM does not expose. The code therefore asserts \texttt{bagging\_fraction}$=1.0$ at runtime.

The only data structure stored after training is the set of \emph{leaf membership vectors} $\boldsymbol{\ell}_{1},\ldots,\boldsymbol{\ell}_{T}$, where $\ell_{t,i}$ is the leaf ID assigned to training instance $i$ in tree $t$.\footnote{For a trained LightGBM model, these are obtained by calling \texttt{predict()} with \texttt{pred\_leaf=True}.} This requires $\mathcal{O}(TN)$ storage and no $N\times N$ matrices are ever formed.

\begin{algorithm}[h!t]
\caption{LeafAverage\label{alg:leaf-average}}
\begin{algorithmic}[1]
\Require $\boldsymbol{v}$ \Comment{$N$-vector}
\Require $\boldsymbol{\ell}$ \Comment{Leaf ID vector for one tree}
\Ensure $\boldsymbol{a}$ \Comment{$N$-vector where each entry is the mean over its leaf}

\Function{LeafAverage}{$\boldsymbol{v},\boldsymbol{\ell}$}
\For{each unique leaf $L$ in $\boldsymbol{\ell}$}
    \State $m \gets \text{mean}\{v_{j}:\ell_{j}=L\}$
    \For{$j$ such that $\ell_{j}=L$}
        \State $a_{j} \gets m$
    \EndFor
\EndFor
\State \textbf{return} $\boldsymbol{a}$
\EndFunction
\end{algorithmic}
\end{algorithm}

Algorithm~\ref{alg:axil-forward} implements the forward operator (Proposition~\ref{prop:forward-operator}). When $\boldsymbol{v}=\boldsymbol{y}$ it reproduces the GBM predictions; for arbitrary $\boldsymbol{v}$ it computes $\boldsymbol{K}\boldsymbol{v}$ without materialising $\boldsymbol{K}$. Each iteration calls \textsc{LeafAverage} once ($\mathcal{O}(N)$), giving $\mathcal{O}(TN)$ total.

\begin{algorithm}[h!t]
\caption{AXIL-Forward\label{alg:axil-forward}}
\begin{algorithmic}[1]
\Require $\boldsymbol{v}$ \Comment{$N$-vector to which $\boldsymbol{K}$ is applied}
\Require $\boldsymbol{\ell}_{1},\ldots,\boldsymbol{\ell}_{T}$ \Comment{Leaf membership vectors}
\Require $\lambda$ \Comment{Learning rate}
\Ensure $\boldsymbol{g}=\boldsymbol{K}\boldsymbol{v}$ \Comment{$N$-vector, cost $\mathcal{O}(TN)$}

\Function{AXIL-Forward}{$\boldsymbol{v},\boldsymbol{\ell}_{1:T},\lambda$}
\State $\boldsymbol{g}\gets\overline{v}\cdot\boldsymbol{1}$ \Comment{Mean of $\boldsymbol{v}$, broadcast to $N$-vector}
\For{$t=1$ to $T$}
    \State $\boldsymbol{g}\gets\boldsymbol{g}+\lambda\cdot\Call{LeafAverage}{\boldsymbol{v}-\boldsymbol{g},\;\boldsymbol{\ell}_{t}}$
\EndFor
\State \textbf{return} $\boldsymbol{g}$
\EndFunction
\end{algorithmic}
\end{algorithm}

Algorithm~\ref{alg:axil-backward} implements the backward operator (Theorem~\ref{thm:backward-operator}). It processes trees in reverse order according to the backward update~(\ref{eq:backward-step}). In the full-batch setting $\boldsymbol{W}_{t}^{T}=\boldsymbol{W}_{t}$ (Appendix~\ref{subsec:W-structural-props}), so each $\boldsymbol{W}_{t}^{T}\boldsymbol{h}_{t}$ reduces to an ordinary \textsc{LeafAverage} call ($\mathcal{O}(N)$), giving $\mathcal{O}(TN)$ total. The AXIL weight vector for prediction~$i$ is row~$i$ of $\boldsymbol{K}$, equal to $\boldsymbol{K}^{T}\boldsymbol{e}_{i}$; setting $\boldsymbol{u}=\boldsymbol{e}_{i}$ therefore returns $\boldsymbol{k}_{i}$ with $\widehat{y}_{i}=\boldsymbol{k}_{i}^{T}\boldsymbol{y}$ exactly. For $S$ predictions, run the operator on $S$ basis vectors (or batch them into a matrix).

\begin{algorithm}[h!t]
\caption{AXIL-Backward\label{alg:axil-backward}}
\begin{algorithmic}[1]
\Require $\boldsymbol{u}$ \Comment{$N$-vector to which $\boldsymbol{K}^{T}$ is applied}
\Require $\boldsymbol{\ell}_{1},\ldots,\boldsymbol{\ell}_{T}$ \Comment{Leaf membership vectors}
\Require $\lambda$ \Comment{Learning rate}
\Ensure $\boldsymbol{k}=\boldsymbol{K}^{T}\boldsymbol{u}$ \Comment{$N$-vector, cost $\mathcal{O}(TN)$}

\Function{AXIL-Backward}{$\boldsymbol{u},\boldsymbol{\ell}_{1:T},\lambda$}
\State $\boldsymbol{h}\gets\boldsymbol{u}$
\State $\boldsymbol{k}\gets\boldsymbol{0}$
\For{$t=T$ down to $1$}
    \State $\boldsymbol{r}\gets\lambda\cdot\Call{LeafAverage}{\boldsymbol{h},\;\boldsymbol{\ell}_{t}}$
    \State $\boldsymbol{k}\gets\boldsymbol{k}+\boldsymbol{r}$
    \State $\boldsymbol{h}\gets\boldsymbol{h}-\boldsymbol{r}$
\EndFor
\State $\boldsymbol{k}\gets\boldsymbol{k}+\overline{h}\cdot\boldsymbol{1}$ \Comment{Add base learner (global mean) contribution}
\State \textbf{return} $\boldsymbol{k}$
\EndFunction
\end{algorithmic}
\end{algorithm}

Algorithm~\ref{alg:axil-oos} implements the out-of-sample backward operator (Theorem~\ref{thm:oos-backward}). For a new instance $\boldsymbol{x}_{\mathrm{new}}$, let $\ell_{t}^{*}$ be the leaf ID assigned to $\boldsymbol{x}_{\mathrm{new}}$ in tree $t$.\footnote{Obtained by calling \texttt{predict()} with \texttt{pred\_leaf=True} on $\boldsymbol{x}_{\mathrm{new}}$.} The cross-leaf vector $\boldsymbol{c}_{t}$ is constructed inside the loop: $c_{t,j}=1/n_{l}$ for training instances $j$ sharing leaf $\ell_{t}^{*}$, and $0$ otherwise. Each step applies \textsc{LeafAverage} and adds the cross-leaf contribution, giving $\mathcal{O}(TN)$ total.

\begin{algorithm}[h!t]
\caption{AXIL-OOS\label{alg:axil-oos}}
\begin{algorithmic}[1]
\Require $\boldsymbol{x}_{\mathrm{new}}$ \Comment{New instance to explain}
\Require $\boldsymbol{\ell}_{1},\ldots,\boldsymbol{\ell}_{T}$ \Comment{Leaf membership vectors (training set)}
\Require $\lambda$ \Comment{Learning rate}
\Ensure $\boldsymbol{k}_{\mathrm{new}}$ \Comment{OOS AXIL weight vector, cost $\mathcal{O}(TN)$}

\Function{AXIL-OOS}{$\boldsymbol{x}_{\mathrm{new}},\boldsymbol{\ell}_{1:T},\lambda$}
\State $\boldsymbol{p}\gets\boldsymbol{0}$ \Comment{$N$-vector}
\For{$t=T$ down to $1$}
    \State $\ell_{t}^{*}\gets$ leaf ID of $\boldsymbol{x}_{\mathrm{new}}$ in tree $t$
    \State $\boldsymbol{c}\gets\boldsymbol{0}$;\quad $n_{l}\gets|\{j:\ell_{t,j}=\ell_{t}^{*}\}|$
    \For{$j$ such that $\ell_{t,j}=\ell_{t}^{*}$}
        \State $c_{j}\gets 1/n_{l}$
    \EndFor
    \State $\boldsymbol{p}\gets\lambda\,\boldsymbol{c}+\boldsymbol{p}-\lambda\cdot\Call{LeafAverage}{\boldsymbol{p},\;\boldsymbol{\ell}_{t}}$
\EndFor
\State $\boldsymbol{k}_{\mathrm{new}}\gets\boldsymbol{p}+\frac{1-\boldsymbol{1}^{T}\boldsymbol{p}}{N}\cdot\boldsymbol{1}$ \Comment{Add base learner contribution}
\State \textbf{return} $\boldsymbol{k}_{\mathrm{new}}$
\EndFunction
\end{algorithmic}
\end{algorithm}

The returned weight vector satisfies $f_{T}(\boldsymbol{x}_{\mathrm{new}})=\boldsymbol{k}_{\mathrm{new}}^{T}\boldsymbol{y}$ exactly.

\paragraph{Computing the full AXIL weight matrix} When the complete $N\times N$ matrix $\boldsymbol{K}$ is needed, apply the backward operator to the $N\times N$ identity: \textsc{LeafAverage} operates independently on each column, and the result is $\boldsymbol{K}^{T}$ directly. Equivalently, the forward operator applied column-by-column to $\boldsymbol{I}$ produces $\boldsymbol{K}$. Both approaches cost $\mathcal{O}(TN^{2})$ and avoid the overhead of $N$ separate function calls.

%======================================================================
\section{Complexity\label{sec:Complexity}}
%======================================================================

\subsection{Time and space complexity}

Table~\ref{tab:complexity-comparison} summarises the time and space complexity of the AXIL algorithms. $N$ is the number of training instances, $T$ is the number of trees, and $S$ is the number of predictions to explain.

\begin{table}[h!t]
\caption{Complexity of AXIL algorithms\label{tab:complexity-comparison}}
\begin{tabular}{lllll}
\textbf{Operation} & \textbf{Algorithm} & \textbf{Time} & \textbf{Space} \tabularnewline
\hline
Post-fit setup (store leaf memberships) & --- & $\mathcal{O}(TN)$ & $\mathcal{O}(TN)$ \tabularnewline
Explain 1 in-sample prediction & AXIL-Backward & $\mathcal{O}(TN)$ & $\mathcal{O}(N)$ \tabularnewline
Explain 1 OOS prediction & AXIL-OOS & $\mathcal{O}(TN)$ & $\mathcal{O}(N)$ \tabularnewline
Explain $S$ predictions & AXIL-Backward $\times S$ & $\mathcal{O}(TNS)$ & $\mathcal{O}(NS)$ \tabularnewline
Full AXIL weight matrix & AXIL-Backward $\times N$ & $\mathcal{O}(TN^{2})$ & $\mathcal{O}(N^{2})$ \tabularnewline
Apply $\boldsymbol{K}$ to a vector & AXIL-Forward & $\mathcal{O}(TN)$ & $\mathcal{O}(N)$ \tabularnewline
Apply $\boldsymbol{K}^{T}$ to a vector & AXIL-Backward & $\mathcal{O}(TN)$ & $\mathcal{O}(N)$ \tabularnewline
\hline
\end{tabular}
\end{table}

All costs follow from the $\mathcal{O}(N)$ cost of applying $\boldsymbol{W}_{t}$ to an $N$-vector (Lemma~\ref{lem:block-diagonal}): every algorithm that iterates over $T$ trees and processes one or more $N$-vectors per tree inherits $\mathcal{O}(TN)$-per-vector scaling. For a fixed number of queried predictions and bounded number of trees, AXIL is therefore asymptotically no slower than just printing the explanations themselves.

\subsection{Comparison with other methods}

Table~\ref{tab:method-comparison} compares AXIL with alternative instance-attribution methods.

\begin{table}[h!t]
\caption{Comparison with alternative instance-attribution methods\label{tab:method-comparison}}
\begin{tabular}{llll}
\textbf{Method} & \textbf{Per-prediction cost} & \textbf{Exact?} & \textbf{Prediction-specific?} \tabularnewline
\hline
AXIL (this paper) & $\mathcal{O}(TN)$ & Yes & Yes \tabularnewline
BoostIn & $\mathcal{O}(TN)$ & No & Yes \tabularnewline
TREX & $\mathcal{O}(N)$ & No & Yes \tabularnewline
LeafInfluence & $\mathcal{O}(N)$ per Hessian-vector product & No & Yes \tabularnewline
DataShapley & $\mathcal{O}(M \cdot \text{retrain})$ & No & No (global) \tabularnewline
\hline
\end{tabular}
\par\smallskip\noindent{\small\emph{Note.} TREX and LeafInfluence per-prediction costs are marginal costs given method-specific setup (kernel fitting for TREX, Hessian computation for LeafInfluence). Table~\ref{tab:timings} reports end-to-end wall-clock times including setup.}
\end{table}

\subsection{Empirical scaling}

We verify the theoretical complexity on synthetic data (Friedman \#1, $T=100$ trees, $L=31$ leaves).\footnote{Generated on a Core i9-14900K processor (24 cores/32 threads), 192\,GB RAM.} Table~\ref{tab:empirical-scaling} reports timings for the full AXIL weight matrix (equivalently, AXIL-Backward applied to the identity matrix, whose columns are the $N$ standard basis vectors) and for $S=10$ predictions. The full matrix is not computed for $N\geq100{,}000$ because it requires $N^{2}\times8$ bytes of memory (80\,GB at $N=100{,}000$).

\begin{table}[h!t]
\caption{Empirical scaling of AXIL\label{tab:empirical-scaling}}
\begin{tabular}{rrr}
\toprule
$N$ & Full matrix (s) & $S{=}10$ queries (s) \\
\midrule
100 & 0.03 & 0.0049 \\
300 & 0.05 & 0.0070 \\
1{,}000 & 0.62 & 0.0087 \\
3{,}000 & 6.55 & 0.02 \\
10{,}000 & 66.1 & 0.08 \\
30{,}000 & 542 & 0.30 \\
100{,}000 & --- & 1.13 \\
300{,}000 & --- & 3.49 \\
1{,}000{,}000 & --- & 11.8 \\
10{,}000{,}000 & --- & 121 \\
\bottomrule
\end{tabular}

\end{table}

Both columns match the predicted scaling: the full-matrix column scales as $\sim N^{2}$ and the $S=10$ column as $\sim N$, with observed multipliers within 10\% of predicted across all steps. At $N=10{,}000{,}000$ with $T=100$ trees, exact AXIL weights for 10 predictions are computed in 121 seconds.

\subsection{Runtime comparison}

Table~\ref{tab:timings} compares the wall-clock time of AXIL against BoostIn, TREX, and LeafInfluence for explaining $S=10$ test predictions on each of the 20 experiment datasets. Starting from an already-fitted GBM, each timing includes the full cost of method-specific explainer setup and computing influence scores for 10 queries.

\begin{table}[h!t]
\caption{Runtime comparison: wall-clock time (seconds) from an already-fitted GBM to explanations for $S=10$ test predictions. Lowest time per dataset in bold.\label{tab:timings}}
\begin{tabular}{l r r r r r}
\toprule
 & $N$ & AXIL & BoostIn & TREX & LeafInf \\
\midrule
Abalone & 4{,}177 & \textbf{0.08} & 0.59 & 1.33 & --- \\
Airfoil & 1{,}503 & \textbf{0.05} & 0.29 & 0.25 & 59.9 \\
Autos & 392 & \textbf{0.03} & 0.18 & 0.14 & 8.05 \\
Bodyfat & 252 & \textbf{0.02} & 0.21 & 0.16 & 4.68 \\
Boston & 506 & \textbf{0.02} & 0.18 & 0.17 & 11.5 \\
Concrete & 1{,}030 & \textbf{0.04} & 0.22 & 0.20 & 32.9 \\
CPU & 8{,}192 & \textbf{0.10} & 1.03 & 7.32 & --- \\
Cars & 804 & \textbf{0.03} & 0.29 & 0.20 & 22.2 \\
Diabetes & 442 & \textbf{0.04} & 0.19 & 0.16 & 9.65 \\
Energy efficiency & 768 & \textbf{0.03} & 0.21 & 0.17 & 21.1 \\
Forest fire & 517 & \textbf{0.03} & 0.22 & 0.17 & 11.7 \\
Grid stability & 10{,}000 & \textbf{0.15} & 1.33 & 1.48 & --- \\
Kin8nm & 8{,}192 & \textbf{0.13} & 0.87 & 1.00 & --- \\
QSAR fish & 908 & \textbf{0.03} & 0.23 & 0.23 & 26.6 \\
Red wine & 1{,}599 & \textbf{0.05} & 0.29 & 0.29 & 64.9 \\
Space ga & 3{,}107 & \textbf{0.07} & 0.49 & 0.48 & --- \\
Titanic & 1{,}307 & \textbf{0.03} & 0.24 & 0.22 & 45.5 \\
Treasury & 1{,}049 & \textbf{0.03} & 0.27 & 0.23 & 34.7 \\
White wine & 4{,}898 & \textbf{0.09} & 0.66 & 0.66 & --- \\
Yacht & 308 & \textbf{0.03} & 0.13 & 0.15 & 5.92 \\
\bottomrule
\end{tabular}

\end{table}

AXIL is the fastest method on every dataset. BoostIn is roughly $4$--$10\times$ slower; TREX is roughly $4$--$75\times$ slower; LeafInfluence, where it runs, is over $100\times$ slower on every dataset and exceeds $1{,}000\times$ on several.

%======================================================================
\section{Experiments\label{sec:experiments}}
%======================================================================

Instance-attribution methods for GBMs assign a score to each training instance for a given prediction, but they differ fundamentally in what that score represents. We first use a target-perturbation experiment to reveal these differences (Section~\ref{subsec:exactness}), then ask whether they matter in practice by measuring faithfulness under retraining (Section~\ref{subsec:faithfulness}). Throughout, we compare against BoostIn, TREX, and LeafInfluence from \citet{brophy2023adapting}, the leading instance-attribution methods for GBMs.

\subsection{Setup\label{subsec:setup}}

We evaluate on 20 standard regression datasets from OpenML \citep{vanschoren2013openml}, ranging from $N=252$ (Bodyfat) to $N=10{,}000$ (Grid stability).\footnote{OpenML IDs: Abalone (45033), Airfoil (44957), Autos (42372), Bodyfat (560), Boston (531), Concrete (44959), CPU (227), Cars (44994), Diabetes (41514), Energy efficiency (44960), Forest fire (43440), Grid stability (44973), Kin8nm (189), QSAR fish (44970), Red wine (44972), Space ga (507), Titanic (41265), Treasury (42367), White wine (44971), Yacht (42370). Categorical variables are one-hot encoded; non-numeric variables and instances with missing values are dropped.} Each dataset is split 80/20 into training and test sets (seed 42). The GBM is a LightGBM regressor \citep{ke2017lightgbm} with $T=100$ trees, 31 leaves, learning rate $\lambda=0.1$, and \texttt{min\_child\_samples}${}=2$. Row subsampling is disabled (\texttt{bagging\_fraction}${}=1.0$, LightGBM's default for \texttt{boosting\_type='gbdt'}) so that all training instances participate in each tree's leaf-averaging step; in the notation of Section~\ref{sec:decomposition}, $\mathcal{C}_{i}^{(t)}=\mathcal{L}_{i}^{(t)}$ for every instance and tree, the full-batch case in which $\boldsymbol{W}_{t}$ is symmetric and doubly stochastic.

All competitor methods are from the \texttt{tree-influence} package \citep{brophy2023adapting}. LeafInfluence is prohibitively slow for $N>2{,}000$ and is omitted on larger datasets.

\subsection{What do the methods actually measure?\label{subsec:exactness}}

AXIL, BoostIn, and TREX all assign a score to each training instance for a given test prediction, but the scores represent different quantities. For any fixed query prediction, AXIL provides exact coefficients in a linear decomposition of that prediction into training targets, so perturbing training target $y_j$ by $\delta$ changes the query prediction by exactly $k_{i,j}\delta$. BoostIn scores measure gradient contributions along the boosting trajectory; TREX scores are weights from an $L2$-regularised kernel surrogate. Neither claims to decompose the prediction itself. Whether this distinction matters in practice depends on how well the competitors track actual target sensitivity; if all methods scored $r > 0.99$, the exactness argument would be largely academic.

To test this, we randomly select $J=30$ training instances and up to 50 test instances as queries. For each selected training instance $j$, we perturb $y_j \to y_j + \delta$ where $\delta = \sigma_y$ (the standard deviation of the training targets). With the tree structure held fixed, the exact change in any query prediction due to this perturbation is the corresponding AXIL weight on instance $j$ times $\delta$. We take this as the ground-truth sensitivity and measure the Pearson correlation $r$ between each method's predicted change and the AXIL-implied change, pooled across all $(j, \text{query})$ pairs. AXIL must achieve $r=1.000$ by construction; the question is whether competitor scores correlate equally well with the true sensitivity.

\begin{table}[h!t]
\caption{Target sensitivity: Pearson $r$ between each method's predicted change and the actual change in test predictions after a training-target perturbation of $\delta=\sigma_y$. A value of 1.000 means the method's scores are exact linear predictors of the perturbation effect.\label{tab:exactness}}
\begin{tabular}{l r r r r r r}
\toprule
 & $N$ & $M$ & AXIL $r$ & BoostIn $r$ & TREX $r$ \\
\midrule
Abalone & 4{,}177 & 7 & \textbf{1.000} & 0.261 & 0.679 \\
Airfoil & 1{,}503 & 5 & \textbf{1.000} & 0.685 & 0.915 \\
Autos & 392 & 5 & \textbf{1.000} & 0.238 & 0.740 \\
Bodyfat & 252 & 14 & \textbf{1.000} & 0.356 & 0.660 \\
Boston & 506 & 22 & \textbf{1.000} & 0.320 & 0.640 \\
Concrete & 1{,}030 & 8 & \textbf{1.000} & 0.036 & 0.810 \\
CPU & 8{,}192 & 12 & \textbf{1.000} & 0.024 & 0.976 \\
Cars & 804 & 17 & \textbf{1.000} & 0.564 & 0.829 \\
Diabetes & 442 & 10 & \textbf{1.000} & 0.218 & 0.532 \\
Energy efficiency & 768 & 8 & \textbf{1.000} & 0.529 & 0.670 \\
Forest fire & 517 & 10 & \textbf{1.000} & 0.131 & 0.133 \\
Grid stability & 10{,}000 & 12 & \textbf{1.000} & 0.252 & 0.163 \\
Kin8nm & 8{,}192 & 8 & \textbf{1.000} & 0.086 & 0.637 \\
QSAR fish & 908 & 6 & \textbf{1.000} & 0.184 & 0.702 \\
Red wine & 1{,}599 & 11 & \textbf{1.000} & 0.211 & 0.794 \\
Space ga & 3{,}107 & 6 & \textbf{1.000} & 0.223 & 0.792 \\
Titanic & 1{,}307 & 7 & \textbf{1.000} & 0.415 & 0.603 \\
Treasury & 1{,}049 & 15 & \textbf{1.000} & 0.195 & 0.826 \\
White wine & 4{,}898 & 11 & \textbf{1.000} & 0.249 & 0.793 \\
Yacht & 308 & 6 & \textbf{1.000} & 0.445 & 0.410 \\
\bottomrule
\end{tabular}

\end{table}

The results are not close. BoostIn averages $r\approx 0.28$: its scores are nearly orthogonal to actual prediction changes, measuring gradient contributions rather than target sensitivity. TREX averages $r\approx 0.67$: it fits a weighted-sum model but uses a kernel surrogate, not the GBM itself. AXIL achieves $r=1.000$ on all 20 datasets, as required. Only AXIL's scores are interpretable as exact sensitivity coefficients: ``the prediction is this weighted sum of training targets.''

\subsection{Faithfulness under retraining\label{subsec:faithfulness}}

AXIL's exactness result holds by construction for the fixed-model target-sensitivity test in Table~\ref{tab:exactness}. A harder question is whether AXIL's rankings remain informative when the model itself changes. We therefore adapt the single-test removal protocol of \citet{brophy2023adapting} to our prediction-level setting. For each of up to 100 randomly selected test predictions (fewer if the test split is smaller than 100 instances), we rank training instances by the absolute magnitude of each method's score for that query, remove the top $\lceil\alpha_{m}N\rceil$ training instances for $\alpha_{m}\in\{0.1\%,\,0.5\%,\,1\%,\,1.5\%,\,2\%\}$, retrain the GBM from scratch, and record the absolute change in that same test prediction. Let $\widehat{y}_{q}$ denote the original prediction for query $q$ and $\widehat{y}_{q}^{-\alpha_{m}}$ the prediction after removal and retraining. The per-query area under the removal curve (AURC) is the mean absolute prediction change across the $M=5$ removal fractions:
\begin{equation}
\mathrm{AURC}_{q}=\frac{1}{M}\sum_{m=1}^{M}\bigl|\widehat{y}_{q}^{-\alpha_{m}}-\widehat{y}_{q}\bigr|\label{eq:aurc}
\end{equation}
Higher AURC means the method's top-ranked instances cause larger prediction changes when removed.

This differs from \citet{brophy2023adapting}'s single-test removal experiment in two ways. First, Brophy \emph{et al.} rank instances by positive influence on the test-example loss and evaluate the increase in that example's loss after retraining. We instead rank by absolute score magnitude and evaluate absolute prediction change. We do this because AXIL explains the prediction itself rather than the loss, and because for prediction-specific attribution both large positive and large negative contributors are influential in the relevant sense: removing either can substantially move the prediction. Retraining changes the tree structure, so this evaluation goes beyond any method's theoretical guarantee. AXIL weights measure fixed-model sensitivity to training targets, not the effect of removing an instance and retraining; AXIL was not designed for this task. Whether AXIL nonetheless identifies the right instances to remove is an empirical question. A Random baseline that ranks instances uniformly at random provides a lower bound for this evaluation.

\begin{table}[h!t]
\caption{Faithfulness under retraining: for each test prediction, training instances are ranked by absolute influence-score magnitude; the top 0.1\%--2\% are removed, the model is retrained, and the absolute change in that same prediction is recorded. The table reports AURC averaged over up to 100 test predictions per dataset. Each dataset occupies two rows: mean AURC (top) and standard errors in parentheses (bottom). Higher is better. \textbf{Bold}: not significantly worse than the best mean in that row (paired two-sided $t$-test; bold if $p\ge 0.05$, i.e.\ no significant difference detected; no multiplicity adjustment). --- indicates the method was too slow. Footer rows: \emph{Clear wins} = sole bold entry; \emph{Incl.\ ties} = bold for any reason.\label{tab:faithfulness}}

\centering
\begingroup
\renewcommand{\arraystretch}{0.82}%
\setlength{\tabcolsep}{3.5pt}%
% Uniform scale so the table fits inside both text width and 70\% of text height
% (height-only resizebox was leaving the table wider than \linewidth).
\adjustbox{center,max width=\linewidth,max height=\dimexpr\textheight*37/100\relax,keepaspectratio}{%
\begin{tabular}{l r r r r r r r}
\toprule
 & $N$ & $M$ & AXIL & BoostIn & TREX & LeafInf & Random \\
\midrule
Abalone & 4{,}177 & 7 & \textbf{0.769} & \textbf{0.691} & \textbf{0.719} & --- & 0.313 \\
 &  &  & {\scriptsize (0.077)} & {\scriptsize (0.070)} & {\scriptsize (0.059)} &  & {\scriptsize (0.029)} \\
Airfoil & 1{,}503 & 5 & \textbf{1.749} & 1.376 & \textbf{1.692} & 1.035 & 0.420 \\
 &  &  & {\scriptsize (0.152)} & {\scriptsize (0.110)} & {\scriptsize (0.159)} & {\scriptsize (0.098)} & {\scriptsize (0.026)} \\
Autos & 392 & 5 & \textbf{1.813} & 1.513 & 1.323 & 1.413 & 0.551 \\
 &  &  & {\scriptsize (0.154)} & {\scriptsize (0.162)} & {\scriptsize (0.117)} & {\scriptsize (0.134)} & {\scriptsize (0.056)} \\
Bodyfat & 252 & 14 & \textbf{0.606} & \textbf{0.560} & 0.504 & \textbf{0.558} & 0.199 \\
 &  &  & {\scriptsize (0.132)} & {\scriptsize (0.123)} & {\scriptsize (0.123)} & {\scriptsize (0.120)} & {\scriptsize (0.039)} \\
Boston & 506 & 22 & \textbf{1.809} & 1.594 & 1.510 & 1.391 & 0.669 \\
 &  &  & {\scriptsize (0.255)} & {\scriptsize (0.264)} & {\scriptsize (0.248)} & {\scriptsize (0.251)} & {\scriptsize (0.075)} \\
Concrete & 1{,}030 & 8 & \textbf{3.842} & 3.269 & 3.387 & 2.545 & 0.852 \\
 &  &  & {\scriptsize (0.263)} & {\scriptsize (0.258)} & {\scriptsize (0.260)} & {\scriptsize (0.215)} & {\scriptsize (0.063)} \\
CPU & 8{,}192 & 12 & \textbf{1.053} & \textbf{1.005} & \textbf{1.011} & --- & 0.430 \\
 &  &  & {\scriptsize (0.105)} & {\scriptsize (0.085)} & {\scriptsize (0.097)} &  & {\scriptsize (0.037)} \\
Cars & 804 & 17 & \textbf{1574.872} & \textbf{1632.416} & \textbf{1739.300} & 1299.793 & 372.616 \\
 &  &  & {\scriptsize (204.165)} & {\scriptsize (287.889)} & {\scriptsize (275.440)} & {\scriptsize (243.309)} & {\scriptsize (44.319)} \\
Diabetes & 442 & 10 & \textbf{22.008} & \textbf{22.934} & 18.704 & 19.206 & 10.759 \\
 &  &  & {\scriptsize (1.608)} & {\scriptsize (1.450)} & {\scriptsize (1.392)} & {\scriptsize (1.341)} & {\scriptsize (0.682)} \\
Energy efficiency & 768 & 8 & 0.359 & \textbf{0.429} & 0.335 & 0.269 & 0.061 \\
 &  &  & {\scriptsize (0.059)} & {\scriptsize (0.071)} & {\scriptsize (0.059)} & {\scriptsize (0.063)} & {\scriptsize (0.004)} \\
Forest fire & 517 & 10 & \textbf{19.430} & \textbf{17.079} & \textbf{17.473} & 16.786 & 5.863 \\
 &  &  & {\scriptsize (2.767)} & {\scriptsize (3.127)} & {\scriptsize (3.163)} & {\scriptsize (3.081)} & {\scriptsize (0.678)} \\
Grid stability & 10{,}000 & 12 & 0.006 & \textbf{0.007} & 0.005 & --- & 0.003 \\
 &  &  & {\scriptsize (0.000)} & {\scriptsize (0.000)} & {\scriptsize (0.000)} &  & {\scriptsize (0.000)} \\
Kin8nm & 8{,}192 & 8 & \textbf{0.077} & \textbf{0.073} & 0.068 & --- & 0.029 \\
 &  &  & {\scriptsize (0.005)} & {\scriptsize (0.005)} & {\scriptsize (0.004)} &  & {\scriptsize (0.001)} \\
QSAR fish & 908 & 6 & \textbf{0.513} & 0.422 & \textbf{0.519} & 0.370 & 0.150 \\
 &  &  & {\scriptsize (0.042)} & {\scriptsize (0.039)} & {\scriptsize (0.047)} & {\scriptsize (0.034)} & {\scriptsize (0.011)} \\
Red wine & 1{,}599 & 11 & \textbf{0.355} & 0.309 & \textbf{0.344} & 0.276 & 0.110 \\
 &  &  & {\scriptsize (0.034)} & {\scriptsize (0.032)} & {\scriptsize (0.032)} & {\scriptsize (0.028)} & {\scriptsize (0.006)} \\
Space ga & 3{,}107 & 6 & \textbf{0.062} & 0.049 & \textbf{0.059} & --- & 0.019 \\
 &  &  & {\scriptsize (0.007)} & {\scriptsize (0.006)} & {\scriptsize (0.006)} &  & {\scriptsize (0.001)} \\
Titanic & 1{,}307 & 7 & \textbf{13.988} & 7.158 & 9.754 & 6.278 & 1.334 \\
 &  &  & {\scriptsize (3.240)} & {\scriptsize (1.504)} & {\scriptsize (2.195)} & {\scriptsize (1.386)} & {\scriptsize (0.182)} \\
Treasury & 1{,}049 & 15 & \textbf{0.118} & 0.101 & 0.107 & 0.078 & 0.030 \\
 &  &  & {\scriptsize (0.014)} & {\scriptsize (0.011)} & {\scriptsize (0.012)} & {\scriptsize (0.012)} & {\scriptsize (0.004)} \\
White wine & 4{,}898 & 11 & \textbf{0.239} & \textbf{0.235} & \textbf{0.241} & --- & 0.102 \\
 &  &  & {\scriptsize (0.021)} & {\scriptsize (0.022)} & {\scriptsize (0.019)} &  & {\scriptsize (0.008)} \\
Yacht & 308 & 6 & \textbf{0.329} & \textbf{0.321} & 0.240 & 0.206 & 0.043 \\
 &  &  & {\scriptsize (0.078)} & {\scriptsize (0.087)} & {\scriptsize (0.059)} & {\scriptsize (0.063)} & {\scriptsize (0.008)} \\
\midrule
\multicolumn{3}{l}{Clear wins} & 5 & 2 & 0 & 0 & 0 \\
\multicolumn{3}{l}{Incl.\ ties} & 18 & 11 & 9 & 1 & 0 \\
\bottomrule
\end{tabular}
}%
\endgroup
\end{table}

Table~\ref{tab:faithfulness} shows that AXIL is statistically indistinguishable from the best method on 18 of 20 datasets (paired $t$-test at $\alpha=0.05$; bold entries), with the highest sample-mean AURC on 14. The footer row \emph{Incl.\ ties} counts every dataset on which that method is bold (clear wins plus shared top): AXIL 18, BoostIn 11, TREX 9. The pattern is asymmetric. The two losses for AXIL (Energy efficiency, Grid stability) share small absolute AURC values, indicating that no method identifies strongly influential instances in these datasets and the different approaches produce rankings of comparable quality. Where instances \emph{are} highly influential (Titanic, Forest fire, Boston), AXIL's exact decomposition provides a clear advantage, with margins well beyond sampling noise and, most notably on Titanic, an AURC of 14.0 versus 9.8 for the next-best method. LeafInfluence, where available ($N\leq 2{,}000$), is the weakest competitor throughout.

\begin{figure}[h!t]
\centering
\includegraphics[width=\textwidth]{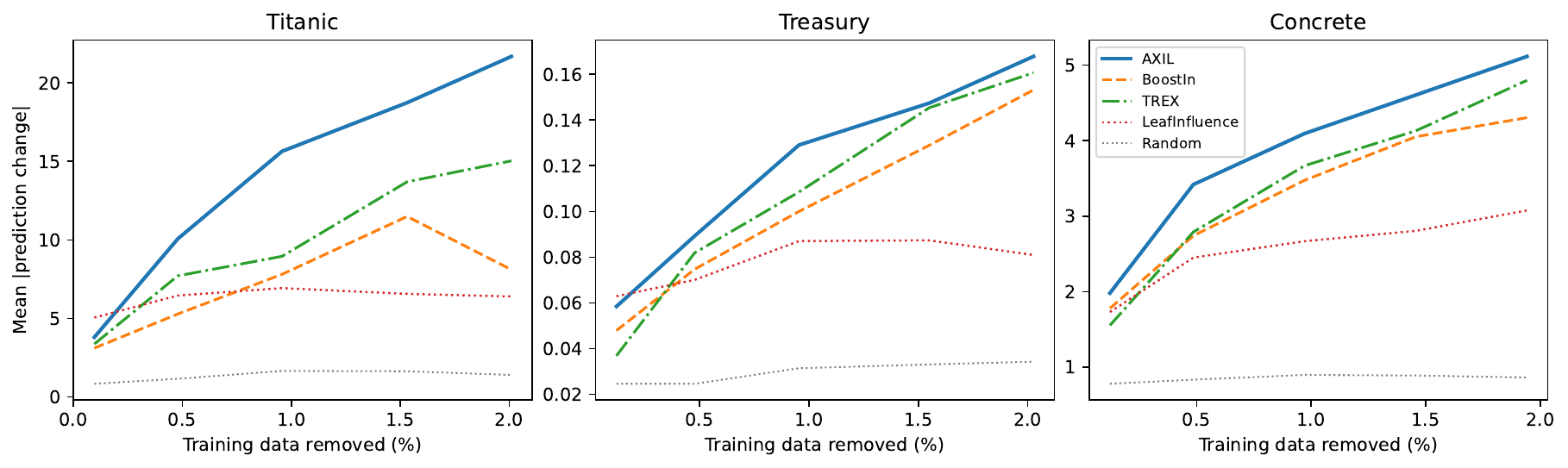}
\caption{Monotone removal curves for three representative datasets. Each curve shows the mean absolute change in test predictions as successively larger fractions of the most influential training instances (according to each method) are removed and the model is retrained. Higher curves indicate more effective identification of influential instances.\label{fig:removal-curves}}
\end{figure}

Figure~\ref{fig:removal-curves} shows the removal curves for three representative datasets: Titanic, Treasury, and Concrete. On Titanic, AXIL's curve rises steeply and separates early from all competitors, indicating that its top-ranked instances are substantially more influential. On Treasury, AXIL remains highest but BoostIn and TREX are closer. On Concrete, AXIL holds a modest lead throughout.

%======================================================================
\section{Boundary Results\label{sec:boundary-results}}
%======================================================================

The AXIL decomposition ($\widehat{\boldsymbol{y}} = \boldsymbol{K}\boldsymbol{y}$) holds whenever the prediction pipeline preserves linearity in the training targets $\boldsymbol{y}$ with the model structure held fixed. It fails when any essential nonlinearity in $\boldsymbol{y}$ enters the pipeline: a nonlinear initial prediction, nonlinear pseudo-residuals, or nonlinear leaf-value computation. This section applies this principle to the main model classes of interest.

The results here share a common structure. Where AXIL holds (Table~\ref{tab:AXIL-applicability}, ``Yes'' rows), we prove it \emph{universally}: every instance of the class admits the decomposition for any training set. For GBM classifiers with log-loss (Section~\ref{subsec:Impossibility-GBM-classification}), we show that on the binary-label domain with nondegenerate training sets and $N\geq 3$, the log-odds initialisation is already nonlinear in $\boldsymbol{y}$, so AXIL fails immediately and can only be restored by exact cancellation at later boosting steps. For neural networks (Section~\ref{subsec:Impossibility-neural-networks}), the class is too broad for such a universal statement at the architecture level, so instead we define a concrete subclass that covers most real-world implementations, Ordinary Neural Networks (ONNs), and show that AXIL is impossible for that class.

\subsection{The linear leaf-update criterion\label{subsec:linear-leaf-update}}

For GBM-type ensembles, the critical requirement is that each tree's leaf-value computation be a linear function of the residual vector.

\begin{prop}[Linear leaf-update criterion]\label{prop:linear-leaf-update}
The AXIL decomposition holds for any GBM variant satisfying two conditions: \emph{(i)} the initial prediction $f_0$ is linear in $\boldsymbol{y}$ (e.g.\ the training-set mean); and \emph{(ii)} at each boosting step, the pseudo-residual vector is linear in $\boldsymbol{y}$ and each tree's per-leaf update is a linear function of that vector. These conditions are sufficient; the GBM-classification impossibility result and the ONN impossibility result in Sections~\ref{subsec:Impossibility-GBM-classification}--\ref{subsec:Impossibility-neural-networks} show that important practical violations of these conditions break the decomposition.
\end{prop}
\begin{proof}
Condition (ii) ensures that the per-tree operator (which may differ from the standard leaf-averaging operator $\boldsymbol{W}_t$) is a fixed linear map on $\mathbb{R}^N$ independent of $\boldsymbol{y}$, so the AXIL recursion (Theorem~\ref{thm:Any-GBM-regression}) applies since its proof requires only linearity of the leaf operators. Condition (i) ensures that the base prediction can be written as $\boldsymbol{K}_0\boldsymbol{y}$ for some fixed matrix $\boldsymbol{K}_0$.
\end{proof}

Standard GBM regression with $L2$ loss satisfies both conditions: the initial prediction is $\bar{y} = \frac{1}{N}\boldsymbol{1}^T\boldsymbol{y}$, and each leaf value is the mean of the residuals in that leaf. $L2$-regularised leaf values (ridge-type shrinkage) also satisfy condition (ii), since the regularised leaf value remains a linear function of the leaf residuals.

The decomposition fails for $L1$ regularisation, which shrinks leaf means toward zero and snaps sufficiently small values to exactly zero (a nonlinear operation), and for quantile-based leaf values, both of which are nonlinear functions of the residuals. More broadly, the restriction to squared-error loss is not a modelling convenience but a structural necessity: for $L2$ loss the pseudo-residuals are $y_i - \widehat{y}_i$, linear in $\boldsymbol{y}$, preserving the linearity chain through the boosting recursion. For other standard losses (absolute error, Huber, quantile) the pseudo-residuals involve nonlinear functions of $\boldsymbol{y}$ such as $\mathrm{sign}(y_i - \widehat{y}_i)$, violating condition (ii) at the first boosting step.

\subsection{Classification trees and Random Forests\label{subsec:Classification-trees-RF}}

For a fitted binary classification tree predicting $y\in\{0,1\}$, with the learned leaf memberships held fixed, the predicted probability for instance $i$ is the proportion of positive labels in the leaf:
\begin{equation}
p_{i}=\frac{1}{N_{i}}\sum_{j\in\mathcal{L}_{i}}y_{j}\label{eq:classification tree}
\end{equation}
which is the same formula as for a regression tree~(\ref{eq:k-tree}). The AXIL decomposition applies with identical weights. For a fitted binary Random Forest classifier, with each tree's leaf memberships held fixed, the predicted probability is the average of the individual tree probabilities, which is linear in $\boldsymbol{y}$ using~(\ref{eq:k-rf}) (again with multiplicity-adjusted coefficients under bootstrap sampling).
\begin{cor}
For any fitted binary classification tree or Random Forest classifier, with the learned tree structure held fixed, each probability prediction is a linear combination of the training targets $\boldsymbol{y}$.\label{cor:classification-trees-RF}
\end{cor}
\begin{proof}
Once the tree structure is fixed,~(\ref{eq:classification tree}) is identical in form to a regression tree prediction~(\ref{eq:k-tree}), so $p_i = \boldsymbol{k}_i^{\text{TREE}}\cdot\boldsymbol{y}$ with the same weights. For a Random Forest, holding the fitted trees fixed gives $\frac{1}{T}\sum_{t=1}^{T}\boldsymbol{k}_i^{\text{TREE},t}\cdot\boldsymbol{y} = \left(\frac{1}{T}\sum_{t=1}^{T}\boldsymbol{k}_i^{\text{TREE},t}\right)\cdot\boldsymbol{y}$, which is linear in $\boldsymbol{y}$.
\end{proof}
For multiclass classification, the same argument applies componentwise to one-hot class-indicator targets, so each class-probability coordinate is linear in the corresponding indicator matrix.

\subsection{Impossibility: GBM classification\label{subsec:Impossibility-GBM-classification}}

For binary GBM classifiers trained with log-loss, the generic obstacle is the initial prediction $g_{0}=\log\bigl(\bar{y}/(1-\bar{y})\bigr)$, the log-odds of the sample proportion. On the binary-label domain this fails to be linear in $\boldsymbol{y}$ once the training set has at least three instances and contains both classes. Linearity therefore breaks at the very first step, violating condition~(i) of Proposition~\ref{prop:linear-leaf-update}. At $T=1$ this already rules out AXIL. For $T\geq 2$, the subsequent pseudo-residuals compound the nonlinearity through the sigmoid function $\sigma(g_{t-1})$, and only an exact cancellation by later boosting terms could restore linearity.
\begin{thm}
For a binary GBM classifier trained with log-loss on a nondegenerate training set with $N\geq 3$, the initial raw score $g_0$ is not a linear combination of the training targets $\boldsymbol{y}$. Consequently, the one-tree raw-score predictions $g_1(\boldsymbol{x})$ are not linear combinations of $\boldsymbol{y}$. More generally, for $T\geq 2$ the raw-score predictions $g_T(\boldsymbol{x})$ remain nonlinear in $\boldsymbol{y}$ unless later boosting terms exactly cancel the nonlinearity introduced by the base logit. Whenever $g_T(\boldsymbol{x})$ is nonlinear and differs from $g_0$ (i.e.\ at least one tree contributes a non-zero update for $\boldsymbol{x}$), the predicted probability $\sigma(g_T(\boldsymbol{x}))$ is nonlinear in $\boldsymbol{y}$ as well.\label{thm:GBM-classification-impossible}
\end{thm}
\begin{proof}
See Appendix~\ref{subsec:proof-gbm-classification}.
\end{proof}

\subsection{Impossibility: Ordinary Neural Networks\label{subsec:Impossibility-neural-networks}}

Neural networks are too broad a class for a sharp universal boundary analogous to GBM classification. The cases that do admit AXIL are essentially degenerate linear-smoother exceptions, most notably identity-map training procedures whose in-sample fitted predictor satisfies $\widehat{\boldsymbol{y}}=\boldsymbol{y}$ for every target vector (hence $\boldsymbol{K}=\boldsymbol{I}$), and fixed-feature linear smoothers, where the hidden representation is held fixed and only a final linear readout is fit by a linear method such as least squares or ridge. We therefore define a concrete class, Ordinary Neural Networks (ONNs), that covers the standard end-to-end cases of practical interest and for which AXIL is impossible.

\begin{defn}[Ordinary Neural Network (ONN)]\label{def:ONN}
Fix a feed-forward regression network trained for $T\geq 1$ gradient-descent steps on squared-error loss. We call the resulting trained model an \emph{Ordinary Neural Network} if there exist a training instance $x_i$ and a scalar prediction component $\widehat{y}_j$ such that, with all targets except $y_i$ held fixed:
\begin{enumerate}
\item \emph{Label-dependent hidden update:} at the first gradient step, changing $y_i$ changes the update of at least one hidden-layer parameter on a trainable path to $\widehat{y}_j$;
\item \emph{Concrete nonlinearity on that path:} along such a path, either
\begin{enumerate}
\item \emph{Smooth case:} some activation on the path is smooth and genuinely curved on an open interval traversed as $y_i$ varies (for example tanh, sigmoid, or GELU); or
\item \emph{Piecewise-linear case:} every activation on the path is piecewise linear but not linear (for example ReLU, leaky-ReLU, or PReLU), all units on the path remain in the same active linear piece as $y_i$ varies in a neighbourhood, and at least two trainable affine layers on the path have first-step updates that change with $y_i$;
\end{enumerate}
\item \emph{No exact cancellation:} after all $T$ steps, the final scalar map $y_i\mapsto \widehat{y}_j$ is not affine on that neighbourhood.
\end{enumerate}
\end{defn}

Condition~1 rules out fixed-feature linear readouts. Condition~2 identifies the two concrete mechanisms that create nonlinearity in the target dependence: a curved activation, or a multilayer product on an active piecewise-linear path. Condition~3 excludes the degenerate case where later training happens to cancel that nonlinearity exactly.

\begin{prop}[AXIL impossible for Ordinary Neural Networks]\label{prop:NN-impossible}
No Ordinary Neural Network admits an AXIL decomposition: for an ONN there is no fixed matrix $\boldsymbol{K}$ such that $\widehat{\boldsymbol{y}}=\boldsymbol{K}\boldsymbol{y}$ for all target vectors $\boldsymbol{y}$.
\end{prop}
\begin{proof}
See Appendix~\ref{subsec:proof-nn-impossible}.
\end{proof}

\subsection{Summary\label{subsec:Summary-table}}

Table~\ref{tab:AXIL-applicability} summarises the applicability of the AXIL decomposition across model classes.

\begin{table}[h!t]
\caption{Applicability of AXIL decomposition across model classes\label{tab:AXIL-applicability}}
{\small\setlength{\tabcolsep}{2pt}
\begin{tabular}{p{0.29\linewidth}p{0.12\linewidth}p{0.19\linewidth}p{0.30\linewidth}}
\textbf{Model class} & \textbf{AXIL?} & \textbf{Reference} & \textbf{Reason}\tabularnewline
\hline
Linear regression & Yes & Sec.~\ref{sec:intro} & Fixed hat matrix\tabularnewline
Regression tree / RF & Yes & Eq.\ (\ref{eq:k-tree}), (\ref{eq:k-rf}) & Linear leaf averaging\tabularnewline
Classification tree / RF & Yes & Cor.~\ref{cor:classification-trees-RF} & Linear class probabilities\tabularnewline
GBM regression ($L2$ loss) & Yes & Thm.~\ref{thm:Any-GBM-regression} & Linear residual updates\tabularnewline
GBM classification (log-loss) & No & Thm.~\ref{thm:GBM-classification-impossible} & Nonlinear base logit\tabularnewline
Ordinary NN (Def.~\ref{def:ONN}) & No & Prop.~\ref{prop:NN-impossible} & Nonlinear target dependence\tabularnewline
\hline
General differentiable learner & Approx. & Prop.~\ref{prop:target-response-jacobian} & First-order via Jacobian\tabularnewline
\hline
\end{tabular}}
\end{table}

\subsection{Beyond exact decomposition: the target-response Jacobian\label{subsec:target-response-jacobian}}

The exact AXIL decomposition requires the prediction map to be globally linear in $\boldsymbol{y}$. While this rules out important model classes such as GBM classifiers and neural networks (Sections~\ref{subsec:Impossibility-GBM-classification}--\ref{subsec:Impossibility-neural-networks}), the core idea -- measuring how predictions respond to training targets -- extends to any differentiable learner as a first-order approximation.

Let $F:\mathbb{R}^{N}\to\mathbb{R}^{S}$ denote the end-to-end map from training targets $\boldsymbol{y}$ to predictions (in-sample or out-of-sample), with features, hyperparameters, and all other inputs held fixed.

\begin{prop}[Target-response Jacobian]\label{prop:target-response-jacobian}
Suppose $F$ is differentiable at $\boldsymbol{y}$, and define the target-response Jacobian $\boldsymbol{J}=D_{\boldsymbol{y}}F(\boldsymbol{y})\in\mathbb{R}^{S\times N}$.
\begin{enumerate}
\item[\emph{(i)}] \emph{First-order attribution.} Each prediction admits the decomposition
\begin{equation}
F(\boldsymbol{y}+\boldsymbol{\delta})=F(\boldsymbol{y})+\boldsymbol{J}\boldsymbol{\delta}+o(\lVert\boldsymbol{\delta}\rVert)\label{eq:jacobian-first-order}
\end{equation}
so entry $J_{qj}=\partial\hat{y}_{q}/\partial y_{j}$ is the first-order analogue of the AXIL weight $k_{q,j}$: row $\boldsymbol{j}_{q}$ of $\boldsymbol{J}$ gives the marginal sensitivity of prediction $q$ to each training target.
\item[\emph{(ii)}] \emph{Exact case.} If $F$ is target-linear, i.e.\ $F(\boldsymbol{y})=\boldsymbol{K}\boldsymbol{y}$ for a fixed matrix $\boldsymbol{K}$ independent of $\boldsymbol{y}$, then $\boldsymbol{J}=\boldsymbol{K}$ identically and the AXIL decomposition is the exact, globally valid special case: the remainder in~(\ref{eq:jacobian-first-order}) vanishes for all $\boldsymbol{\delta}$.
\item[\emph{(iii)}] \emph{Implicit-differentiation formula.} Suppose $F(\boldsymbol{y})=h(\theta^{*}(\boldsymbol{y}))$ where $h$ is differentiable and $\theta^{*}(\boldsymbol{y})=\arg\min_{\theta}\mathcal{L}(\theta,\boldsymbol{y})$ for a twice continuously differentiable loss $\mathcal{L}$ whose parameter Hessian $\boldsymbol{\mathcal{H}}=\nabla_{\theta}^{2}\mathcal{L}(\theta^{*},\boldsymbol{y})$ is positive definite. Then
\begin{equation}
\boldsymbol{J}=-\boldsymbol{G}\,\boldsymbol{\mathcal{H}}^{-1}\boldsymbol{C}\label{eq:jacobian-implicit}
\end{equation}
where $\boldsymbol{G}=\partial h/\partial\theta\big|_{\theta^{*}}\in\mathbb{R}^{S\times P}$ maps parameters to predictions and $\boldsymbol{C}=\partial^{2}\mathcal{L}/\partial\theta\,\partial\boldsymbol{y}^{T}\big|_{\theta^{*}}\in\mathbb{R}^{P\times N}$ is the loss cross-derivative.
\end{enumerate}
\end{prop}
\begin{proof}
Part~(i) is the definition of differentiability. Part~(ii) follows because the derivative of a linear map is the map itself: $D_{\boldsymbol{y}}(\boldsymbol{K}\boldsymbol{y})=\boldsymbol{K}$. For~(iii), differentiate the first-order optimality condition $\nabla_{\theta}\mathcal{L}(\theta^{*}(\boldsymbol{y}),\boldsymbol{y})=\boldsymbol{0}$ with respect to $\boldsymbol{y}$: the chain rule gives $\boldsymbol{\mathcal{H}}\,\partial\theta^{*}/\partial\boldsymbol{y}+\boldsymbol{C}=\boldsymbol{0}$. Since $\boldsymbol{\mathcal{H}}\succ 0$, we have $\partial\theta^{*}/\partial\boldsymbol{y}=-\boldsymbol{\mathcal{H}}^{-1}\boldsymbol{C}$, and the chain rule applied to $h$ yields $\boldsymbol{J}=\boldsymbol{G}(-\boldsymbol{\mathcal{H}}^{-1}\boldsymbol{C})$.
\end{proof}

Proposition~\ref{prop:target-response-jacobian} places the AXIL weight matrix $\boldsymbol{K}$ inside a broader framework: $\boldsymbol{K}$ is the globally constant special case of the target-response Jacobian $\boldsymbol{J}$. For any twice-differentiable parametric learner -- including neural networks trained with weight decay and kernel machines -- the Jacobian is computable via~(\ref{eq:jacobian-implicit}) using standard implicit-differentiation techniques \citep{lorraine2020optimizing,franceschi2018bilevel,koh2017understanding}. The rows of $\boldsymbol{J}$ provide local, prediction-specific instance attribution analogous to AXIL weights: $J_{qj}$ measures how much prediction $q$ would change if training target $y_j$ were perturbed by a small amount. Whether this first-order approximation is faithful for general learners, and how to compute it efficiently at scale, are questions left for future work; this paper develops the exact case where the approximation is unnecessary.

%======================================================================
\section{Conclusion}
%======================================================================

We present AXIL, an exact instance-attribution method for squared-error GBMs centred on a matrix-free backward operator. Building on the linear-smoother structure of $L2$ boosting, we show that each fitted prediction can be written as a weighted sum of training targets, which yields exact, prediction-specific instance weights for both in-sample and out-of-sample predictions. The backward operator computes one such weight vector in $\mathcal{O}(TN)$ time, linear in the training set size, without forming any $N\times N$ matrix. The exact decomposition extends to classification trees and Random Forests but is provably impossible for log-loss GBM classifiers and standard neural networks (Table~\ref{tab:AXIL-applicability}).

AXIL applies to Gradient Boosting Machines (GBMs), one of the dominant supervised learning methods for tabular data. This is a setting where instance-level attribution is particularly actionable, as individual training records remain inherently meaningful and interpretable. The exactness guarantee means that AXIL weights are not approximations to be validated but ground-truth sensitivities of the fitted predictor to its training targets. In experiments on 20 regression datasets, this translates into the best faithfulness score on 14 datasets and statistical ties for best on a further 4, while running faster than all competing methods.

Where the exact decomposition does not hold, the target-response Jacobian~(\ref{eq:jacobian-implicit}) provides a principled first-order analogue: the AXIL weight matrix is the special case where this Jacobian is constant. Investigating the quality of the Jacobian approximation for general learners, and developing efficient algorithms for computing it, are natural directions for future work.

%======================================================================
% Acknowledgements, Funding, and Competing Interests (JMLR required)
%======================================================================

\acks{Both authors contributed equally to all aspects of the paper.
Code is available at \url{https://github.com/pgeertsema/AXIL_paper} under the GNU General Public License.
Datasets are available from OpenML (\url{https://openml.org}).

\textbf{Funding:} This research was not funded by any third party.

\textbf{Competing interests:} There are no conflicts of interest to disclose.

\textbf{Generative AI:} During the preparation of this work the authors used large language models (Claude, ChatGPT, Gemini) to assist with literature search, checking theoretical results and code, and proofreading. The authors reviewed and edited all content and take full responsibility for the publication.}

\newpage{}

\bibliography{refs}

\begin{thebibliography}{29}
\providecommand{\natexlab}[1]{#1}
\providecommand{\url}[1]{\texttt{#1}}
\expandafter\ifx\csname urlstyle\endcsname\relax
  \providecommand{\doi}[1]{doi: #1}\else
  \providecommand{\doi}{doi: \begingroup \urlstyle{rm}\Url}\fi

\bibitem[Adadi and Berrada(2018)]{adadi2018peeking}
Amina Adadi and Mohammed Berrada.
\newblock Peeking inside the black-box: a survey on explainable artificial intelligence (xai).
\newblock \emph{IEEE access}, 6:\penalty0 52138--52160, 2018.

\bibitem[Arrieta et~al.(2020)Arrieta, D{\'\i}az-Rodr{\'\i}guez, Del~Ser, Bennetot, Tabik, Barbado, Garc{\'\i}a, Gil-L{\'o}pez, Molina, Benjamins, et~al.]{arrieta2020explainable}
Alejandro~Barredo Arrieta, Natalia D{\'\i}az-Rodr{\'\i}guez, Javier Del~Ser, Adrien Bennetot, Siham Tabik, Alberto Barbado, Salvador Garc{\'\i}a, Sergio Gil-L{\'o}pez, Daniel Molina, Richard Benjamins, et~al.
\newblock Explainable artificial intelligence (xai): Concepts, taxonomies, opportunities and challenges toward responsible ai.
\newblock \emph{Information fusion}, 58:\penalty0 82--115, 2020.

\bibitem[Borisov et~al.(2021)Borisov, Leemann, Se{\ss}ler, Haug, Pawelczyk, and Kasneci]{borisov2021deep}
Vadim Borisov, Tobias Leemann, Kathrin Se{\ss}ler, Johannes Haug, Martin Pawelczyk, and Gjergji Kasneci.
\newblock Deep neural networks and tabular data: {A} survey.
\newblock \emph{CoRR}, abs/2110.01889, 2021.
\newblock URL \url{https://arxiv.org/abs/2110.01889}.

\bibitem[Breiman(2001)]{breiman2001random}
Leo Breiman.
\newblock Random forests.
\newblock \emph{Machine learning}, 45:\penalty0 5--32, 2001.

\bibitem[Brophy et~al.(2023)Brophy, Hammoudeh, and Lowd]{brophy2023adapting}
Jonathan Brophy, Zayd Hammoudeh, and Daniel Lowd.
\newblock Adapting and evaluating influence-estimation methods for gradient-boosted decision trees.
\newblock \emph{Journal of Machine Learning Research}, 24\penalty0 (154):\penalty0 1--48, 2023.

\bibitem[B{\"u}hlmann and Yu(2003)]{buhlmann2003boosting}
Peter B{\"u}hlmann and Bin Yu.
\newblock Boosting with the {$L_2$} loss: Regression and classification.
\newblock \emph{Journal of the American Statistical Association}, 98\penalty0 (462):\penalty0 324--339, 2003.
\newblock \doi{10.1198/016214503000125}.

\bibitem[Franceschi et~al.(2018)Franceschi, Frasconi, Salzo, Grazzi, and Pontil]{franceschi2018bilevel}
Luca Franceschi, Paolo Frasconi, Saverio Salzo, Riccardo Grazzi, and Massimiliano Pontil.
\newblock Bilevel programming for hyperparameter optimization and meta-learning.
\newblock In \emph{Proceedings of the 35th International Conference on Machine Learning}, volume~80 of \emph{PMLR}, pages 1568--1577, 2018.

\bibitem[Ghorbani and Zou(2019)]{ghorbani2019data}
Amirata Ghorbani and James Zou.
\newblock Data shapley: Equitable valuation of data for machine learning.
\newblock In \emph{International Conference on Machine Learning}, pages 2242--2251. PMLR, 2019.

\bibitem[Grinsztajn et~al.(2022)Grinsztajn, Oyallon, and Varoquaux]{grinsztajn2022why}
L{\'e}o Grinsztajn, Edouard Oyallon, and Ga{\"e}l Varoquaux.
\newblock Why do tree-based models still outperform deep learning on tabular data?
\newblock \emph{arXiv preprint arXiv:2207.08815}, 2022.

\bibitem[Gunning et~al.(2019)Gunning, Stefik, Choi, Miller, Stumpf, and Yang]{gunning2019xai}
David Gunning, Mark Stefik, Jaesik Choi, Timothy Miller, Simone Stumpf, and Guang-Zhong Yang.
\newblock Xai-explainable artificial intelligence.
\newblock \emph{Science robotics}, 4\penalty0 (37):\penalty0 eaay7120, 2019.

\bibitem[Hastie et~al.(2009)Hastie, Tibshirani, and Friedman]{hastie2009elements}
Trevor Hastie, Robert Tibshirani, and Jerome Friedman.
\newblock \emph{The Elements of Statistical Learning}.
\newblock Springer, 2nd edition edition, 2009.

\bibitem[Hoaglin and Welsch(1978)]{hoaglin1978hat}
David~C. Hoaglin and Roy~E. Welsch.
\newblock The hat matrix in regression and {ANOVA}.
\newblock \emph{The American Statistician}, 32\penalty0 (1):\penalty0 17--22, 1978.

\bibitem[Ilyas et~al.(2022)Ilyas, Park, Engstrom, Leclerc, and Madry]{ilyas2022datamodels}
Andrew Ilyas, Sung~Min Park, Logan Engstrom, Guillaume Leclerc, and Aleksander Madry.
\newblock Datamodels: Predicting predictions from training data.
\newblock In \emph{Proceedings of the 39th International Conference on Machine Learning}, volume 162 of \emph{PMLR}, pages 9525--9587, 2022.

\bibitem[Kamath and Liu(2021)]{kamath2021explainable}
Uday Kamath and John Liu.
\newblock \emph{Explainable Artificial Intelligence: An Introduction to Interpretable Machine Learning}.
\newblock Springer, 2021.

\bibitem[Ke et~al.(2017)Ke, Meng, Finley, Wang, Chen, Ma, Ye, and Liu]{ke2017lightgbm}
Guolin Ke, Qi~Meng, Thomas Finley, Taifeng Wang, Wei Chen, Weidong Ma, Qiwei Ye, and Tie-Yan Liu.
\newblock {LightGBM}: A highly efficient gradient boosting decision tree.
\newblock In \emph{Advances in Neural Information Processing Systems}, pages 3146--3154, 2017.

\bibitem[Koh and Liang(2017)]{koh2017understanding}
Pang~Wei Koh and Percy Liang.
\newblock Understanding black-box predictions via influence functions.
\newblock In \emph{International Conference on Machine Learning}, pages 1885--1894. PMLR, 2017.

\bibitem[Linardatos et~al.(2020)Linardatos, Papastefanopoulos, and Kotsiantis]{linardatos2020explainable}
Pantelis Linardatos, Vasilis Papastefanopoulos, and Sotiris Kotsiantis.
\newblock Explainable ai: A review of machine learning interpretability methods.
\newblock \emph{Entropy}, 23\penalty0 (1):\penalty0 18, 2020.

\bibitem[Lorraine et~al.(2020)Lorraine, Vicol, and Duvenaud]{lorraine2020optimizing}
Jonathan Lorraine, Paul Vicol, and David Duvenaud.
\newblock Optimizing millions of hyperparameters by implicit differentiation.
\newblock In \emph{Proceedings of the Twenty Third International Conference on Artificial Intelligence and Statistics}, volume 108 of \emph{PMLR}, pages 1540--1552, 2020.

\bibitem[Lundberg and Lee(2017)]{lundberg2017unified}
Scott~M Lundberg and Su-In Lee.
\newblock A unified approach to interpreting model predictions.
\newblock In \emph{Advances in Neural Information Processing Systems}, pages 4765--4774, 2017.

\bibitem[Lundberg et~al.(2020)Lundberg, Erion, Chen, DeGrave, Prutkin, Nair, Katz, Himmelfarb, Bansal, and Lee]{lundberg2020local}
Scott~M Lundberg, Gabriel Erion, Hugh Chen, Alex DeGrave, Jordan~M Prutkin, Bala Nair, Ronit Katz, Jonathan Himmelfarb, Nisha Bansal, and Su-In Lee.
\newblock From local explanations to global understanding with explainable ai for trees.
\newblock \emph{Nature Machine Intelligence}, 2\penalty0 (1):\penalty0 56--67, 2020.
\newblock \doi{10.1038/s42256-019-0138-9}.

\bibitem[Molnar(2020)]{molnar2020interpretable}
Christoph Molnar.
\newblock \emph{Interpretable machine learning}.
\newblock 2020.

\bibitem[Park et~al.(2023)Park, Georgiev, Ilyas, Leclerc, and Madry]{park2023trak}
Sung~Min Park, Kristian Georgiev, Andrew Ilyas, Guillaume Leclerc, and Aleksander Madry.
\newblock {TRAK}: Attributing model behavior at scale.
\newblock In \emph{Proceedings of the 40th International Conference on Machine Learning}, volume 202 of \emph{PMLR}, pages 27074--27113, 2023.

\bibitem[Pruthi et~al.(2020)Pruthi, Liu, Kale, and Sundararajan]{pruthi2020estimating}
Garima Pruthi, Frederick Liu, Satyen Kale, and Mukund Sundararajan.
\newblock Estimating training data influence by tracing gradient descent.
\newblock In \emph{Advances in Neural Information Processing Systems}, volume~33, pages 19920--19930, 2020.

\bibitem[Ribeiro et~al.(2016)Ribeiro, Singh, and Guestrin]{ribeiro2016why}
Marco~Tulio Ribeiro, Sameer Singh, and Carlos Guestrin.
\newblock "why should i trust you?" explaining the predictions of any classifier.
\newblock In \emph{Proceedings of the 22nd ACM SIGKDD international conference on knowledge discovery and data mining}, pages 1135--1144, 2016.

\bibitem[Scornet(2016)]{scornet2016random}
Erwan Scornet.
\newblock Random forests and kernel methods.
\newblock \emph{IEEE Transactions on Information Theory}, 62\penalty0 (3):\penalty0 1485--1500, 2016.
\newblock \doi{10.1109/TIT.2016.2514489}.

\bibitem[Sharchilev et~al.(2018)Sharchilev, Ustinovskiy, Serdyukov, and de~Rijke]{sharchilev2018finding}
Boris Sharchilev, Yury Ustinovskiy, Pavel Serdyukov, and Maarten de~Rijke.
\newblock Finding influential training samples for gradient boosted decision trees.
\newblock In \emph{Proceedings of the 35th International Conference on Machine Learning}, volume~80, pages 4577--4585. PMLR, 2018.

\bibitem[Vanschoren et~al.(2013)Vanschoren, van Rijn, Bischl, and Torgo]{vanschoren2013openml}
Joaquin Vanschoren, Jan~N. van Rijn, Bernd Bischl, and Luis Torgo.
\newblock Openml: networked science in machine learning.
\newblock \emph{SIGKDD Explorations}, 15\penalty0 (2):\penalty0 49--60, 2013.
\newblock \doi{10.1145/2641190.2641198}.
\newblock URL \url{http://doi.acm.org/10.1145/2641190.264119}.

\bibitem[Xu et~al.(2019)Xu, Uszkoreit, Du, Fan, Zhao, and Zhu]{xu2019explainable}
Feiyu Xu, Hans Uszkoreit, Yangzhou Du, Wei Fan, Dongyan Zhao, and Jun Zhu.
\newblock Explainable ai: A brief survey on history, research areas, approaches and challenges.
\newblock In \emph{CCF international conference on natural language processing and Chinese computing}, pages 563--574. Springer, 2019.

\bibitem[Yeh et~al.(2018)Yeh, Kim, Yen, and Ravikumar]{yeh2018representer}
Chih-Kuan Yeh, Joon Kim, Ian En-Hsu Yen, and Pradeep~K Ravikumar.
\newblock Representer point selection for explaining deep neural networks.
\newblock In \emph{Advances in Neural Information Processing Systems}, volume~31, 2018.

\end{thebibliography}

\appendix

\section{Structural properties of \texorpdfstring{$\boldsymbol{W}_t$}{W\_t}}\label{subsec:W-structural-props}

The main text defers the proof of Lemma~\ref{lem:block-diagonal}. We give that proof first, then record fuller structural properties of $\boldsymbol{W}_t$.

\paragraph{Proof of Lemma~\ref{lem:block-diagonal}.}
Once the ensemble is fitted, the leaf assignment of tree $t$ is fixed; under row subsampling the associated contributing sets are fixed as well. The operator $\boldsymbol{W}_{t}$ therefore depends only on this fitted combinatorial structure, not on the vector it is applied to. In the full-batch setting, linearity is immediate from~(\ref{eq:W-def}); under row subsampling, the same formula holds with $\mathcal{L}_{i}^{(t)}$ replaced by $\mathcal{C}_{i}^{(t)}$, so the map remains linear.

For $\boldsymbol{W}_{t}\boldsymbol{v}$, one pass over the $N$ instances accumulates the relevant sums and counts by leaf, and a second pass writes the leaf mean back to each entry of the output vector. This reads and writes only $\mathcal{O}(N)$ numbers, so the total cost is $\mathcal{O}(N)$.

In the full-batch setting, $\boldsymbol{W}_{t}^{T}=\boldsymbol{W}_{t}$, so the same $\mathcal{O}(N)$ bound holds for the transposed application. Under row subsampling, each contributing index in a leaf receives the same value, namely the leaf sum divided by the size of the contributing set. Concretely, if leaf $l$ has full leaf $\mathcal{L}_{l}^{(t)}$ and contributing set $\mathcal{C}_{l}^{(t)}$, then
\begin{equation}
(\boldsymbol{W}_{t}^{T}\boldsymbol{h})_{j}=
\begin{cases}
\frac{1}{|\mathcal{C}_{l}^{(t)}|}\sum_{k\in\mathcal{L}_{l}^{(t)}} h_{k}, & j\in\mathcal{C}_{l}^{(t)} \text{ for some leaf } l,\\[6pt]
0, & j\notin\mathcal{S}_{t}.
\end{cases}
\end{equation}
Thus one pass over the $N$ instances computes the leaf sums, and a second pass writes the scaled leaf sum to the contributing indices. This costs $\sum_{l}|\mathcal{L}_{l}^{(t)}|+\sum_{l}|\mathcal{C}_{l}^{(t)}|=N+|\mathcal{S}_{t}|\leq 2N=\mathcal{O}(N)$. \qed

\begin{prop}[Full structural properties of $\boldsymbol{W}_t$]\label{prop:W-structure}
For any tree $t$, the leaf-averaging operator $\boldsymbol{W}_{t}$ is block-diagonal with one block per leaf (in the ordering induced by the leaf partition; in arbitrary instance order it is permutation-similar to a block-diagonal matrix). Each block for leaf $l$ (with $n_l = |\mathcal{L}_l|$ instances in full batch) has the form $\frac{1}{n_{l}}\boldsymbol{1}\boldsymbol{1}^{T}\in\mathbb{R}^{n_{l}\times n_{l}}$. $\boldsymbol{W}_{t}$ is idempotent ($\boldsymbol{W}_{t}^{2}=\boldsymbol{W}_{t}$) and row-stochastic. In the full-batch setting, $\boldsymbol{W}_{t}$ is additionally symmetric and doubly stochastic. Applying $\boldsymbol{W}_{t}$ or $\boldsymbol{W}_{t}^{T}$ to an $N\times N$ matrix costs $\mathcal{O}(N^{2})$.
\end{prop}
\begin{proof}
\emph{The fixed-linearity and $\mathcal{O}(N)$ vector-application claims are exactly Lemma~\ref{lem:block-diagonal}.} It remains to prove the additional structural properties and the $\mathcal{O}(N^{2})$ matrix-application cost.

\emph{Block-diagonal structure.} The leaf partition of tree $t$ depends only on the feature splits applied to fixed instances; it is independent of $\boldsymbol{K}_{t-1}$ or any previous tree. Let $\mathcal{C}_{l}^{(t)}=\mathcal{S}_{t}\cap\mathcal{L}_{l}$ denote the \emph{contributing set} for leaf $l$ in tree $t$, where $\mathcal{S}_{t}\subseteq\{1,\ldots,N\}$ is the set of instances that participate in tree $t$'s training (with $\mathcal{S}_{t}=\{1,\ldots,N\}$ in full batch). Within each leaf $l$, all instances share the same contributing set $\mathcal{C}_{l}^{(t)}$ (with $m_l = |\mathcal{C}_{l}^{(t)}|$ and $n_l = |\mathcal{L}_l|$), so cross-leaf entries of $\boldsymbol{W}_t$ are zero. The block for leaf $l$ has every row equal to $\frac{1}{m_{l}}\boldsymbol{e}_{\mathcal{C}_{l}^{(t)}}^{T}$, where $\boldsymbol{e}_{\mathcal{C}_{l}^{(t)}}$ is the indicator of $\mathcal{C}_{l}^{(t)}$ within $\mathcal{L}_{l}$. In the full-batch case ($m_l = n_l$), this reduces to $\frac{1}{n_{l}}\boldsymbol{1}\boldsymbol{1}^{T}\in\mathbb{R}^{n_{l}\times n_{l}}$.

\emph{Idempotence.} Each block has the form $B_{l}=\boldsymbol{1}_{n_{l}}\boldsymbol{c}_{l}^{T}$ where $\boldsymbol{c}_{l}=\frac{1}{m_{l}}\boldsymbol{e}_{\mathcal{C}_{l}^{(t)}}$. Then $B_{l}^{2}=\boldsymbol{1}_{n_{l}}(\boldsymbol{c}_{l}^{T}\boldsymbol{1}_{n_{l}})\boldsymbol{c}_{l}^{T}=\boldsymbol{1}_{n_{l}}\boldsymbol{c}_{l}^{T}=B_{l}$, since $\boldsymbol{c}_{l}^{T}\boldsymbol{1}_{n_{l}}=m_{l}/m_{l}=1$.

\emph{Row-stochasticity.} Every row of block $l$ sums to $m_{l}\cdot(1/m_{l})=1$.

\emph{Symmetry (full-batch).} When $m_{l}=n_{l}$, $\boldsymbol{c}_{l}=\frac{1}{n_{l}}\boldsymbol{1}_{n_{l}}$, so $B_{l}=\frac{1}{n_{l}}\boldsymbol{1}_{n_{l}}\boldsymbol{1}_{n_{l}}^{T}$, which is symmetric.

\emph{Matrix-application cost.} For $\boldsymbol{W}_{t}\boldsymbol{M}$ ($\boldsymbol{M}\in\mathbb{R}^{N\times N}$): replacing the $n_{l}$ rows of leaf $l$ with their contributing-set mean costs $n_{l}N$ per leaf; summing over leaves gives $\mathcal{O}(N^{2})$. For $\boldsymbol{W}_{t}^{T}\boldsymbol{M}$: row $j$ of the result is non-zero only for $j\in\mathcal{C}_{l}^{(t)}$, where it equals $\frac{1}{m_{l}}\sum_{k\in\mathcal{L}_{l}}\boldsymbol{M}_{k,\cdot}$; computing each leaf row-sum costs $\mathcal{O}(n_{l}N)$ and copying to the $m_{l}$ contributing rows costs $\mathcal{O}(m_{l}N)$, giving $\sum_{l}(n_{l}+m_{l})N\leq 2N^{2}=\mathcal{O}(N^{2})$.
\end{proof}

\section{Proof of Theorem~\ref{thm:GBM-classification-impossible}}\label{subsec:proof-gbm-classification}

Consider a GBM classifier trained with log-loss on binary targets $\boldsymbol{y}\in\{0,1\}^{N}$, with $N\geq 3$ and both classes present. The initial prediction (in log-odds space) is:
\begin{equation}\label{eq:gbm-class-g0}
g_{0}=\log\left(\frac{\bar{y}}{1-\bar{y}}\right)
\end{equation}
where $\bar{y}=\frac{1}{N}\sum_{j=1}^{N}y_{j}$. Let $m=\sum_{j=1}^{N}y_j$, so $m\in\{1,\ldots,N-1\}$ by nondegeneracy and
\begin{equation}
g_0=\log\left(\frac{m}{N-m}\right).
\end{equation}
Suppose, for contradiction, that $g_{0}=\boldsymbol{k}\cdot\boldsymbol{y}$ for some fixed weight vector $\boldsymbol{k}\in\mathbb{R}^{N}$. Because $g_0$ depends on $\boldsymbol{y}$ only through the class count $m$ and is invariant under permutations of the labels, comparing binary vectors with a single positive in different positions shows that all components of $\boldsymbol{k}$ must be equal: $k_j=c$ for some constant $c$. Hence $g_0=cm$ for every $m\in\{1,\ldots,N-1\}$.

Now evaluate at the two valid nondegenerate class counts $m=1$ and $m=N-1$:
\begin{equation}
c = g_0(1)=\log\left(\frac{1}{N-1}\right)=-\log(N-1),
\end{equation}
while
\begin{equation}
c(N-1)=g_0(N-1)=\log\left(\frac{N-1}{1}\right)=\log(N-1),
\qquad
c=\frac{\log(N-1)}{N-1}.
\end{equation}
For $N\geq 3$ these two values of $c$ are unequal, a contradiction. Therefore $g_0$ is not linear in $\boldsymbol{y}$ on the binary-label domain once $N\geq 3$ and both classes are present.

In particular, for $T=1$ the prediction for instance $i$ is
\begin{equation}\label{eq:gbm-class-T1}
g_{1}(x_{i}) = \underbrace{\log\!\left(\frac{\bar{y}}{1-\bar{y}}\right)}_{\text{nonlinear in }\boldsymbol{y}} + \lambda\underbrace{\bigl(\bar{y}_{\mathcal{L}_{i}}-\bar{y}\bigr)}_{\text{linear in }\boldsymbol{y}}
\end{equation}
where $\bar{y}_{\mathcal{L}_{i}}$ is the mean of $y_{j}$ over instances in the same leaf as $x_{i}$ (using $\sigma(g_{0})=\bar{y}$; the tree contribution shown uses the gradient-only leaf value; Newton-step implementations such as LightGBM scale by the inverse Hessian $1/[\bar{y}(1-\bar{y})]$, but this does not affect the argument since the base term $g_{0}$ is already nonlinear). Adding a linear function of $\boldsymbol{y}$ to a nonlinear function yields a nonlinear function, so $g_{1}$ is not linear in $\boldsymbol{y}$. For $T\geq 2$ each subsequent pseudo-residual $r_{i}^{(t)}=y_{i}-\sigma(g_{t-1}(x_{i}))$ again involves a nonlinear function of $\boldsymbol{y}$ through $\sigma$, so the nonlinearity is preserved at every boosting step. Finally, when $g_T(\boldsymbol{x})\neq g_0$ for some instance $\boldsymbol{x}$, the raw score is a non-trivial nonlinear function of $\boldsymbol{y}$, and composing with the strictly monotone $\sigma$ preserves this nonlinearity (note that $\sigma(g_0)=\bar{y}$ is linear, so the degenerate case $g_T=g_0$ is excluded). \qed

\section{Proof of Proposition~\ref{prop:NN-impossible}}\label{subsec:proof-nn-impossible}

Let the ONN satisfy Definition~\ref{def:ONN} for training instance $x_i$ and scalar prediction component $\widehat{y}_j$. Hold all targets except $y_i$ fixed.

Under squared-error loss, the gradient of any parameter at the first step is linear in the residuals, hence affine in $y_i$. By Condition~1, at least one hidden-layer parameter on the chosen path has a first-step update that changes with $y_i$, so after the first step that parameter is a non-constant affine function of $y_i$.

If Condition~2(a) holds, this non-constant affine dependence is fed through an activation $\phi$ with $\phi''\neq 0$ on an open interval. The composition $\phi(a+by_i)$ with $b\neq 0$ is not affine in $y_i$, so the corresponding contribution to $\widehat{y}_j$ is non-affine.

If Condition~2(b) holds, all activations on the chosen path remain in the same active linear piece, so on that neighbourhood the path behaves like a composition of affine maps. Since at least two trainable affine layers on the path have first-step updates that change with $y_i$, the resulting contribution to $\widehat{y}_j$ contains a product of two non-constant affine functions of $y_i$, hence a non-zero quadratic term. It is therefore not affine in $y_i$.

Thus Conditions~1 and~2 identify a concrete first-step mechanism that introduces non-affine dependence of $\widehat{y}_j$ on $y_i$. Condition~3 states that this non-affine dependence is not cancelled exactly by the remaining $T-1$ gradient steps, so the final scalar map $y_i\mapsto \widehat{y}_j$ is not affine.

Suppose for contradiction that an AXIL decomposition exists: $\widehat{\boldsymbol{y}}=\boldsymbol{K}\boldsymbol{y}$ for some fixed matrix $\boldsymbol{K}$. Then the $j$-th prediction has the form
\begin{equation}
\widehat{y}_j=\sum_{m=1}^{N}k_{j,m}y_m.
\end{equation}
Holding all targets except $y_i$ fixed, this becomes
\begin{equation}
\widehat{y}_j = k_{j,i}y_i + c
\end{equation}
for some constant $c$, which is affine in $y_i$. This contradicts Condition~3. Therefore no such fixed matrix $\boldsymbol{K}$ exists. \qed

\section{Conservation of influence}\label{subsec:conservation}

\begin{prop}[Mean-preservation and conservation of influence]\label{prop:conservation}
Every row of $\boldsymbol{K}$ sums to one: $\boldsymbol{K}\boldsymbol{1}=\boldsymbol{1}$. In the full-batch setting, every column also sums to one: $\boldsymbol{1}^{T}\boldsymbol{K}=\boldsymbol{1}^{T}$. Both properties hold equally for regression trees and Random Forests.
\end{prop}
\begin{proof}
\emph{Row sums.}\; $\boldsymbol{K}_{0}\boldsymbol{1}=\boldsymbol{1}$ by definition. If $\boldsymbol{K}_{t-1}\boldsymbol{1}=\boldsymbol{1}$, then $(\boldsymbol{I}-\boldsymbol{K}_{t-1})\boldsymbol{1}=\boldsymbol{0}$, so $\boldsymbol{K}_{t}\boldsymbol{1}=\boldsymbol{K}_{t-1}\boldsymbol{1}+\lambda\boldsymbol{W}_{t}\cdot\boldsymbol{0}=\boldsymbol{1}$ by~(\ref{eq:Gt-recursion}).

\emph{Column sums (full-batch).}\; In the full-batch setting $\boldsymbol{W}_{t}$ is symmetric (Appendix~\ref{subsec:W-structural-props}), so $\boldsymbol{1}^{T}\boldsymbol{W}_{t}=(\boldsymbol{W}_{t}\boldsymbol{1})^{T}=\boldsymbol{1}^{T}$. Then $\boldsymbol{1}^{T}\boldsymbol{K}_{0}=\boldsymbol{1}^{T}$, and if $\boldsymbol{1}^{T}\boldsymbol{K}_{t-1}=\boldsymbol{1}^{T}$, left-multiplying~(\ref{eq:Gt-recursion}) by $\boldsymbol{1}^{T}$ gives $\boldsymbol{1}^{T}\boldsymbol{K}_{t}=\boldsymbol{1}^{T}\boldsymbol{K}_{t-1}+\lambda\boldsymbol{1}^{T}\boldsymbol{W}_{t}(\boldsymbol{I}-\boldsymbol{K}_{t-1})=\boldsymbol{1}^{T}+\lambda\boldsymbol{1}^{T}(\boldsymbol{I}-\boldsymbol{K}_{t-1})=\boldsymbol{1}^{T}$.

For a single tree the weight matrix is $\boldsymbol{W}_{t}$ itself, which is row-stochastic and (in full batch) doubly stochastic. For a Random Forest the weight matrix is $\frac{1}{T}\sum_{t}\boldsymbol{W}_{t}$, which inherits both properties in the full-batch setting; under bootstrap sampling the multiplicity-adjusted forest weights remain row-stochastic, but need not be doubly stochastic.
\end{proof}

\section{The forward operator}\label{subsec:forward-operator-appendix}

Because the recursion~(\ref{eq:Gt-recursion}) is linear, applying it to a \emph{vector} $\boldsymbol{v}\in\mathbb{R}^{N}$ instead of the full identity matrix gives $\boldsymbol{K}\boldsymbol{v}$ directly, without ever building $\boldsymbol{K}$.

\begin{prop}[Forward operator]\label{prop:forward-operator}
For any $\boldsymbol{v}\in\mathbb{R}^{N}$, define:
\begin{align}
\boldsymbol{g}_{0} &= \overline{v}\cdot\boldsymbol{1}\nonumber\\
\boldsymbol{g}_{t} &= \boldsymbol{g}_{t-1}+\lambda\,\boldsymbol{W}_{t}(\boldsymbol{v}-\boldsymbol{g}_{t-1}),\quad t=1,\ldots,T\label{eq:forward-operator}
\end{align}
Then $\boldsymbol{g}_{T}=\boldsymbol{K}\boldsymbol{v}$. Each step applies $\boldsymbol{W}_{t}$ to an $N$-vector, costing $\mathcal{O}(N)$ (Lemma~\ref{lem:block-diagonal}), giving $\mathcal{O}(TN)$ total.
\end{prop}
\begin{proof}
By induction on $t$. Let $P(t)$ be the claim $\boldsymbol{g}_{t}=\boldsymbol{K}_{t}\boldsymbol{v}$.
The base case $P(0)$ holds: $\boldsymbol{g}_{0}=\overline{v}\cdot\boldsymbol{1}=\frac{1}{N}\boldsymbol{1}\boldsymbol{1}^{T}\boldsymbol{v}=\boldsymbol{K}_{0}\boldsymbol{v}$.
For the inductive step, assume $\boldsymbol{g}_{t-1}=\boldsymbol{K}_{t-1}\boldsymbol{v}$. Then by~(\ref{eq:forward-operator}):
\begin{equation}
\boldsymbol{g}_{t}
=\boldsymbol{K}_{t-1}\boldsymbol{v}+\lambda\boldsymbol{W}_{t}(\boldsymbol{v}-\boldsymbol{K}_{t-1}\boldsymbol{v})
=\bigl[\boldsymbol{K}_{t-1}+\lambda\boldsymbol{W}_{t}(\boldsymbol{I}-\boldsymbol{K}_{t-1})\bigr]\boldsymbol{v}
=\boldsymbol{K}_{t}\boldsymbol{v}
\end{equation}
where the last equality uses the $\boldsymbol{K}$ recursion~(\ref{eq:Gt-recursion}).
\end{proof}

\section{Worked example (\texorpdfstring{$N=4$, $T=2$, $\lambda=\tfrac{1}{2}$}{N=4, T=2, λ=1/2})\label{subsec:worked-example}}

This section traces the AXIL recursion, forward operator, and backward operator on a minimal example. The reader can verify every entry by hand.

\paragraph{Setup.} Four training instances, two trees, learning rate $\lambda=\tfrac{1}{2}$. Tree~1 has leaves $\{1,2\}$ and $\{3,4\}$; tree~2 has leaves $\{1,3\}$ and $\{2,4\}$. Instance~1 shares a leaf with instance~2 in tree~1 only, with instance~3 in tree~2 only, but does not share a leaf with instance~4 in any tree. (These two partitions happen to satisfy $\boldsymbol{W}_1\boldsymbol{W}_2=\boldsymbol{W}_2\boldsymbol{W}_1$, so $\boldsymbol{K}$ is symmetric here; generically it is not.) The leaf-averaging matrices are:
\begin{equation}\label{eq:worked-W}
\boldsymbol{W}_1 = \tfrac{1}{2}\begin{pmatrix}1&1&0&0\\1&1&0&0\\0&0&1&1\\0&0&1&1\end{pmatrix},
\qquad
\boldsymbol{W}_2 = \tfrac{1}{2}\begin{pmatrix}1&0&1&0\\0&1&0&1\\1&0&1&0\\0&1&0&1\end{pmatrix}
\end{equation}

\paragraph{AXIL recursion (computing $\boldsymbol{K}$ explicitly).}

\emph{Step~0.}\; $\boldsymbol{K}_0 = \tfrac{1}{4}\boldsymbol{1}\boldsymbol{1}^{T}$: every prediction is the global mean with weight $\tfrac{1}{4}$ on each target.

\emph{Step~1.}\; The unexplained weight operator is:
\begin{equation}
\boldsymbol{I}-\boldsymbol{K}_0 =
\begin{pmatrix}
 \tfrac{3}{4}  & -\tfrac{1}{4} & -\tfrac{1}{4} & -\tfrac{1}{4} \\[2pt]
-\tfrac{1}{4}  &  \tfrac{3}{4} & -\tfrac{1}{4} & -\tfrac{1}{4} \\[2pt]
-\tfrac{1}{4}  & -\tfrac{1}{4} &  \tfrac{3}{4} & -\tfrac{1}{4} \\[2pt]
-\tfrac{1}{4}  & -\tfrac{1}{4} & -\tfrac{1}{4} &  \tfrac{3}{4}
\end{pmatrix}
\end{equation}
Applying $\boldsymbol{W}_1$ averages rows pairwise (rows $\{1,2\}$ together, rows $\{3,4\}$ together):
\begin{equation}
\boldsymbol{W}_1(\boldsymbol{I}-\boldsymbol{K}_0) =
\begin{pmatrix}
 \tfrac{1}{4}  &  \tfrac{1}{4}  & -\tfrac{1}{4} & -\tfrac{1}{4} \\[2pt]
 \tfrac{1}{4}  &  \tfrac{1}{4}  & -\tfrac{1}{4} & -\tfrac{1}{4} \\[2pt]
-\tfrac{1}{4}  & -\tfrac{1}{4}  &  \tfrac{1}{4} &  \tfrac{1}{4} \\[2pt]
-\tfrac{1}{4}  & -\tfrac{1}{4}  &  \tfrac{1}{4} &  \tfrac{1}{4}
\end{pmatrix}
\end{equation}
\begin{equation}
\boldsymbol{K}_1 = \boldsymbol{K}_0 + \tfrac{1}{2}\boldsymbol{W}_1(\boldsymbol{I}-\boldsymbol{K}_0) =
\begin{pmatrix}
\tfrac{3}{8} & \tfrac{3}{8} & \tfrac{1}{8} & \tfrac{1}{8} \\[2pt]
\tfrac{3}{8} & \tfrac{3}{8} & \tfrac{1}{8} & \tfrac{1}{8} \\[2pt]
\tfrac{1}{8} & \tfrac{1}{8} & \tfrac{3}{8} & \tfrac{3}{8} \\[2pt]
\tfrac{1}{8} & \tfrac{1}{8} & \tfrac{3}{8} & \tfrac{3}{8}
\end{pmatrix}
\end{equation}
After one tree, co-leaf pairs $\{1,2\}$ and $\{3,4\}$ carry weight $\tfrac{3}{8}$; cross-leaf pairs carry $\tfrac{1}{8}$. Note that $\boldsymbol{K}_1$ is symmetric (this holds for a single tree; see Section~\ref{subsec:comp-challenge}).

\emph{Step~2.}\; The unexplained weight operator and tree~2's contribution:
\begin{equation}
\boldsymbol{I}-\boldsymbol{K}_1 =
\begin{pmatrix}
 \tfrac{5}{8}  & -\tfrac{3}{8} & -\tfrac{1}{8} & -\tfrac{1}{8} \\[2pt]
-\tfrac{3}{8}  &  \tfrac{5}{8} & -\tfrac{1}{8} & -\tfrac{1}{8} \\[2pt]
-\tfrac{1}{8}  & -\tfrac{1}{8} &  \tfrac{5}{8} & -\tfrac{3}{8} \\[2pt]
-\tfrac{1}{8}  & -\tfrac{1}{8} & -\tfrac{3}{8} &  \tfrac{5}{8}
\end{pmatrix}
\end{equation}
Applying $\boldsymbol{W}_2$ averages the tree-2 leaf groups ($\{1,3\}$ and $\{2,4\}$):
\begin{equation}
\boldsymbol{W}_2(\boldsymbol{I}-\boldsymbol{K}_1) =
\begin{pmatrix}
 \tfrac{1}{4}  & -\tfrac{1}{4} &  \tfrac{1}{4} & -\tfrac{1}{4} \\[2pt]
-\tfrac{1}{4}  &  \tfrac{1}{4} & -\tfrac{1}{4} &  \tfrac{1}{4} \\[2pt]
 \tfrac{1}{4}  & -\tfrac{1}{4} &  \tfrac{1}{4} & -\tfrac{1}{4} \\[2pt]
-\tfrac{1}{4}  &  \tfrac{1}{4} & -\tfrac{1}{4} &  \tfrac{1}{4}
\end{pmatrix}
\end{equation}
\begin{equation}
\boldsymbol{K} = \boldsymbol{K}_2 = \boldsymbol{K}_1 + \tfrac{1}{2}\boldsymbol{W}_2(\boldsymbol{I}-\boldsymbol{K}_1) =
\begin{pmatrix}
\tfrac{1}{2} & \tfrac{1}{4} & \tfrac{1}{4} & 0 \\[2pt]
\tfrac{1}{4} & \tfrac{1}{2} & 0 & \tfrac{1}{4} \\[2pt]
\tfrac{1}{4} & 0 & \tfrac{1}{2} & \tfrac{1}{4} \\[2pt]
0 & \tfrac{1}{4} & \tfrac{1}{4} & \tfrac{1}{2}
\end{pmatrix}
\end{equation}
Reading row~1: $\widehat{y}_1 = \tfrac{1}{2}y_1 + \tfrac{1}{4}y_2 + \tfrac{1}{4}y_3 + 0\cdot y_4$. Instance~1 receives weight $\tfrac{1}{2}$ on itself; $\tfrac{1}{4}$ on instance~2 (co-leaf in tree~1 only); $\tfrac{1}{4}$ on instance~3 (co-leaf in tree~2 only); and $0$ on instance~4 (shares no leaf in either tree). All row sums and column sums equal~$1$, as guaranteed by Proposition~\ref{prop:conservation}.

\paragraph{Verification that $\widehat{\boldsymbol{y}}=\boldsymbol{K}\boldsymbol{y}$.} Setting $\boldsymbol{y}=[1,2,3,4]^{T}$:
\begin{equation}
\boldsymbol{K}\boldsymbol{y}=
\begin{pmatrix}\tfrac{1}{2}&\tfrac{1}{4}&\tfrac{1}{4}&0\\[2pt]\tfrac{1}{4}&\tfrac{1}{2}&0&\tfrac{1}{4}\\[2pt]\tfrac{1}{4}&0&\tfrac{1}{2}&\tfrac{1}{4}\\[2pt]0&\tfrac{1}{4}&\tfrac{1}{4}&\tfrac{1}{2}\end{pmatrix}
\begin{pmatrix}1\\2\\3\\4\end{pmatrix}
=
\begin{pmatrix}\tfrac{1}{2}+\tfrac{2}{4}+\tfrac{3}{4}+0\\[2pt]\tfrac{1}{4}+1+0+1\\[2pt]\tfrac{1}{4}+0+\tfrac{3}{2}+1\\[2pt]0+\tfrac{1}{2}+\tfrac{3}{4}+2\end{pmatrix}
=
\begin{pmatrix}\tfrac{7}{4}\\[2pt]\tfrac{9}{4}\\[2pt]\tfrac{11}{4}\\[2pt]\tfrac{13}{4}\end{pmatrix}
=\widehat{\boldsymbol{y}}.
\end{equation}

\paragraph{Backward operator (extracting row~1 without forming $\boldsymbol{K}$).}

We set $\boldsymbol{u}=\boldsymbol{e}_1=[1,0,0,0]^{T}$ to extract $\boldsymbol{k}_1=\boldsymbol{K}^{T}\boldsymbol{e}_1$ (row~1 of $\boldsymbol{K}$).

\emph{Initialise:}\; $\boldsymbol{h}=[1,\,0,\,0,\,0]^{T}$;\quad $\boldsymbol{k}=[0,\,0,\,0,\,0]^{T}$

\emph{Step $t=2$ (tree~2, processed first because backward).}\; Leaf-average $\boldsymbol{h}$ under tree~2: leaf $\{1,3\}$ mean $=\tfrac{1}{2}(1+0)=\tfrac{1}{2}$; leaf $\{2,4\}$ mean $=\tfrac{1}{2}(0+0)=0$.
\begin{align}
\boldsymbol{r} &= \tfrac{1}{2}\cdot[\tfrac{1}{2},\,0,\,\tfrac{1}{2},\,0]^{T} = [\tfrac{1}{4},\,0,\,\tfrac{1}{4},\,0]^{T}\\
\boldsymbol{k} &\leftarrow [0,0,0,0]^{T} + [\tfrac{1}{4},0,\tfrac{1}{4},0]^{T} = [\tfrac{1}{4},\,0,\,\tfrac{1}{4},\,0]^{T}\\
\boldsymbol{h} &\leftarrow [1,0,0,0]^{T} - [\tfrac{1}{4},0,\tfrac{1}{4},0]^{T} = [\tfrac{3}{4},\,0,\,-\tfrac{1}{4},\,0]^{T}
\end{align}

\emph{Step $t=1$ (tree~1).}\; Leaf-average $\boldsymbol{h}$ under tree~1: leaf $\{1,2\}$ mean $=\tfrac{1}{2}(\tfrac{3}{4}+0)=\tfrac{3}{8}$; leaf $\{3,4\}$ mean $=\tfrac{1}{2}(-\tfrac{1}{4}+0)=-\tfrac{1}{8}$.
\begin{align}
\boldsymbol{r} &= \tfrac{1}{2}\cdot[\tfrac{3}{8},\,\tfrac{3}{8},\,-\tfrac{1}{8},\,-\tfrac{1}{8}]^{T} = [\tfrac{3}{16},\,\tfrac{3}{16},\,-\tfrac{1}{16},\,-\tfrac{1}{16}]^{T}\\
\boldsymbol{k} &\leftarrow [\tfrac{1}{4},0,\tfrac{1}{4},0]^{T} + [\tfrac{3}{16},\tfrac{3}{16},-\tfrac{1}{16},-\tfrac{1}{16}]^{T} = [\tfrac{7}{16},\,\tfrac{3}{16},\,\tfrac{3}{16},\,-\tfrac{1}{16}]^{T}\\
\boldsymbol{h} &\leftarrow [\tfrac{3}{4},0,-\tfrac{1}{4},0]^{T} - [\tfrac{3}{16},\tfrac{3}{16},-\tfrac{1}{16},-\tfrac{1}{16}]^{T} = [\tfrac{9}{16},\,-\tfrac{3}{16},\,-\tfrac{3}{16},\,\tfrac{1}{16}]^{T}
\end{align}

\emph{Add base-learner term:}\; $\overline{h_0} = \tfrac{1}{4}(\tfrac{9}{16}-\tfrac{3}{16}-\tfrac{3}{16}+\tfrac{1}{16}) = \tfrac{1}{4}\cdot\tfrac{4}{16} = \tfrac{1}{16}$.
\begin{equation}\label{eq:backward-example-result}
\boldsymbol{k}_1 = [\tfrac{7}{16},\,\tfrac{3}{16},\,\tfrac{3}{16},\,-\tfrac{1}{16}]^{T} + [\tfrac{1}{16},\,\tfrac{1}{16},\,\tfrac{1}{16},\,\tfrac{1}{16}]^{T} = [\tfrac{1}{2},\,\tfrac{1}{4},\,\tfrac{1}{4},\,0]^{T}
\end{equation}
This matches row~1 of $\boldsymbol{K}$ computed above, confirming the backward operator without forming any $4\times 4$ matrix.

\end{document}